\pdfoutput=1

\documentclass[11pt]{article}

\usepackage[final]{acl}

\usepackage{times}
\usepackage{latexsym}

\usepackage[T1]{fontenc}

\usepackage[utf8]{inputenc}

\usepackage{microtype}

\usepackage{inconsolata}

\usepackage{graphicx}

\usepackage{url}
\usepackage{CJKutf8}

\usepackage{multirow}
\usepackage{booktabs,arydshln}
\usepackage{xcolor}
\usepackage{amsmath}
\usepackage{amsfonts}
\usepackage{algorithm}
\usepackage{algorithmic}
\usepackage{subfigure}
\usepackage{tabu}
\usepackage{amssymb}
\usepackage{makecell}

\makeatletter
\def\adl@drawiv#1#2#3{%
        \hskip.5\tabcolsep
        \xleaders#3{#2.5\@tempdimb #1{1}#2.5\@tempdimb}%
                #2\z@ plus1fil minus1fil\relax
        \hskip.5\tabcolsep}
\newcommand{\cdashlinelr}[1]{%
  \noalign{\vskip\aboverulesep
           \global\let\@dashdrawstore\adl@draw
           \global\let\adl@draw\adl@drawiv}
  \cdashline{#1}
  \noalign{\global\let\adl@draw\@dashdrawstore
           \vskip\belowrulesep}}
\makeatother

\newcommand{\rom}[1]{\uppercase\expandafter{\romannumeral #1\relax}}

\title{TokAlign: Efficient Vocabulary Adaptation via Token Alignment}

\author{Chong Li, Jiajun Zhang\footnotemark[1], Chengqing Zong \\
        State Key Laboratory of Multimodal Artificial Intelligence Systems, \\
        Institute of Automation, CAS, Beijing, China\\
        School of Artificial Intelligence, University of Chinese Academy of Sciences, Beijing, China\\
        lichong2021@ia.ac.cn, \{jjzhang, cqzong\}@nlpr.ia.ac.cn
        }

\begin{document}
\maketitle

\renewcommand{\thefootnote}{\fnsymbol{footnote}} 
\footnotetext[1]{Corresponding author.}

\renewcommand{\thefootnote}{\arabic{footnote}}

\begin{abstract}
Tokenization serves as a foundational step for Large Language Models (LLMs) to process text. 
In new domains or languages, the inefficiency of the tokenizer will slow down the training and generation of LLM. 
The mismatch in vocabulary also hinders deep knowledge transfer between LLMs like token-level distillation. 
To mitigate this gap, we propose an efficient method named {\bf TokAlign} to replace the vocabulary of LLM from the token co-occurrences view, and further transfer the token-level knowledge between models. 
It first aligns the source vocabulary to the target one by learning a one-to-one mapping matrix for token IDs. 
Model parameters, including embeddings, are rearranged and progressively fine-tuned for the new vocabulary. 
Our method significantly improves multilingual text compression rates and vocabulary initialization for LLMs, decreasing the perplexity from 3.4$\text{e}^2$ of strong baseline methods to 1.2$\text{e}^2$ after initialization. 
Experimental results on models across multiple parameter scales demonstrate the effectiveness and generalization of TokAlign, which costs as few as 5k steps to restore the performance of the vanilla model. 
After unifying vocabularies between LLMs, token-level distillation can remarkably boost (+4.4\% than sentence-level distillation) the base model, costing only 235M tokens. 
\footnote{Our codes and model are available at \href{https://github.com/ZNLP/TokAlign}{https://github.com/\\ZNLP/TokAlign}}
\end{abstract}



\section{Introduction}
\begin{figure}[t]
\centering
\includegraphics[width=0.45\textwidth]{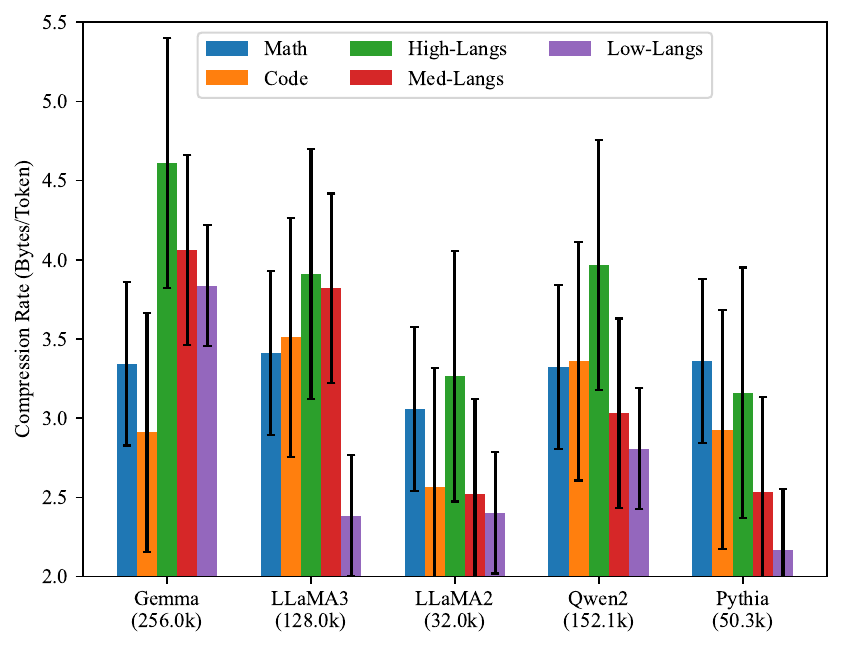}\vspace{-2mm}
\vspace{-1mm}
\caption{The compression rates of tokenizers across different domains and languages, which are still low in the code domain and low-resource languages for most of tokenizers. Refer to Table \ref{tab:tok} in Appendix \ref{appendix:tok} for more details.}
\label{fig:compression}
\vspace{-3mm}
\end{figure}

Large language models \citep{touvron2023llama, openai2023gpt4, yang2024qwen2} first tokenize text input into several tokens during inference and training, which compresses text and addresses the out-of-vocabulary problem \citep{sennrich-etal-2016-neural, wu2016wordpieces, kudo-2018-subword}. 
However, the low compression rate of vanilla tokenizers on new languages or domains decelerates the training and inference process. 
As shown in Figure \ref{fig:compression}, the compression rate of capable large language models like LLaMA3 \citep{meta2024llama3} on low-resource languages still largely lags behind the others. 
For example, Armenian text is 3.95x longer in tokens than English text under the same byte size with the LLaMA3 tokenizer. 
On the other hand, each LLM has specific strengths and weaknesses, which arise from its pre-training corpus and method. 
The mismatch in the vocabulary impedes the deep knowledge transfer between them like token-level distillation and ensemble \citep{xu-etal-2024-bridging, lu2024survey}. 
Considering the huge cost of re-training LLM for a new tokenizer, it is important to investigate efficient vocabulary adaptation methods. 


To address the problems above, we introduce a novel method called {\bf TokAlign} for large language models from a view of token-token co-occurrences. 
It is motivated by the general process of training an LLM: the pre-training corpus is first tokenized into tokens, and then input into the model. 
Given the same pre-training corpus, different tokenizers result in various sequences of token IDs, while the semantic and syntactic information is preserved in the token-token co-occurrence. 
Therefore, TokAlign strives to align token IDs from the original vocabulary and the target ones based on the global token-token co-occurrence matrix \citep{pennington-etal-2014-glove} and learns a token-token alignment matrix. 
We further propose two metrics to evaluate the performance of the token-token alignment matrix based on text matching and semantic similarity. 
Given the learned alignment matrix, the new target embedding and language modeling head of LLM (``$lm\_head$'' in the Transformers \citep{wolf2019huggingface}) are initialized from the parameters of the most similar source token. 
Further vocabulary adaptation process is divided into a progressive two-stage procedure to improve the stability of convergence. 

Given a target multilingual vocabulary for substitution, the model trained on the English corpus obtains a good initialization, decreasing the perplexity from 3.4e${}^{2}$ to 1.2e${}^{2}$, and improves 29.2\% compression rates across 13 languages on average. 
The training process of TokAlign is 1.92x faster than strong baseline methods, and does not require additional hundreds of GPU hours to train a hyper-network for embedding initialization \citep{minixhofer2024zett}. 
Experimental results on models across different scales show that as few as 5k steps are needed for our method to recover the performance of vanilla models on the general domain. 
Moreover, unifying vocabulary between models further facilitates the token-level distillation, which is 4.4\% better than the sentence-level distillation on the same corpus. 
The performance of the 1B model is comparable with the vanilla 7B model after token-level distillation from a capable LLM. 
In summary, our contributions are as follows:
\begin{itemize}
    \item We propose an unsupervised method to align token IDs between two vocabularies and replace the vocabulary of LLMs from the token-token co-occurrence view. 
    \item We introduce two metrics to evaluate the performance of the token-level alignment matrix learned, which are proportional to the initial loss of pre-training. 
    \item Experimental results on ten datasets show that our method promotes the cross-lingual knowledge transfer among multiple languages and deep knowledge transfer between models like token-level distillation.
\end{itemize}


\section{Related Works}
Our work is related to word representation, large language models, and vocabulary adaption, which will be briefly introduced below. 

\paragraph{Word Representation} Based on the distributional semantic hypothesis, \citet{bengio2003neural} introduced the neural probabilistic language model to learn word representation. 
Researchers mainly focus on improving the effectiveness during learning word representations \citep{mikolov2013cbow, mikolov2013skip, bojanowski2017fasttext, li2017implicit, wang2018learning}, which provide a good initialization for neural networks like LSTM and GRU \citep{hochreiter1997lstm, chung2014empirical}. 
GloVe \citep{pennington-etal-2014-glove} provides a method to train word representations from a view of global word-word co-occurrence matrix decomposition. 
It motivates us to train a word representation for each token and align tokens from statistical co-occurrence information in the pre-training corpus. 

\paragraph{Large Language Model} Through scaling in the parameters and pre-training corpus \citep{kaplan2020scaling, hoffmann2022chinchilla}, large language models like GPT-4 and LLaMA3 \citep{radford2018gpt, radford2019gpt2, brown-et-al-2020-gpt3, openai2023gpt4, touvron2023llama, touvron2023llama2, meta2024llama3, glm2024chatglm} demonstrate impressive performance across multiple tasks. 
However, the mismatch in the vocabulary greatly hinders the deep knowledge transfer between different models. 
We aim to mitigate this problem by introducing an efficient method to replace the tokenizer of a large language model. 

\begin{figure*}[t]
\centering
\includegraphics [width=0.9\textwidth]{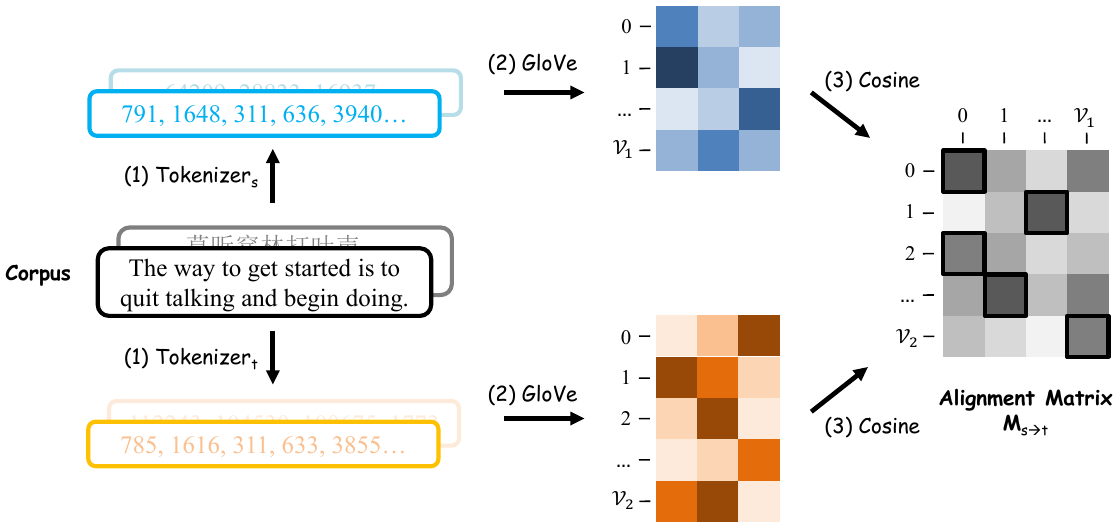}
\vspace{-2mm}
\caption{\label{fig:method} Illustration of TokAlign to align token IDs from different vocabularies. We train token representations on the tokenized corpus, and align token IDs by the cosine similarity. It is noted that the IDs of tokens belonging to both vocabularies are directly replaced without alignment.}
\vspace{-2mm}
\end{figure*}

\paragraph{Vocabulary Adaption} is investigated mainly in the multilingual domain, especially the cross-lingual knowledge transfer problem \citep{le2023bloom, muennighoff-etal-2023-crosslingual, yang2023bigtranslate, zhu2023extrapolating, ustun-etal-2024-aya, li-etal-2024-improving-context, liu-etal-2024-ofa, minixhofer2024zett, yamaguchi-etal-2024-empirical, mundra-etal-2024-empirical, blade2024MEDVOC}. 
It aims to improve the encoding effectiveness of tokenizer on corpora from new languages or domains, and is often implemented by extending the original vocabulary \citep{tran2020english, chau-etal-2020-parsing, minixhofer-etal-2022-wechsel, dobler-de-melo-2023-focus, downey-etal-2023-embedding}. 
Most methods, like Focus \citep{dobler-de-melo-2023-focus}, rely on the tokens belonging to both source vocabulary and target vocabulary to initialize the other new tokens in the target vocabulary. 
Our method differs from these studies for the whole replacement of vocabulary and does not rely on the tokens in both source vocabulary and target vocabulary.

The pipeline of TokAlign to adapt vocabulary is similar to WECHSEL\citep{minixhofer-etal-2022-wechsel}, while the main difference lies in the representation and alignment of tokens. 
WECHSEL requires a bilingual dictionary and word representation to align tokens and calculates the similarity between tokens by tokenizing all words in the dictionary and linearly composing word representations. 
In contrast, TokAlign conducts token representation learning and alignment in an unsupervised way, which can apply to languages without bilingual dictionaries. 

\section{Method: TokAlign}
\subsection{Vocabulary Alignment}
\label{sec:vocab_align}
As shown in Figure \ref{fig:method}, there are three steps for TokAlign to align two vocabularies from the token-token co-occurrence information. 
We denote the source tokenizer as $\text{Tokenizer}_{s}$, which has $\mathcal{V}_s$ tokens, and the target tokenizer as $\text{Tokenizer}_{t}$ with $\mathcal{V}_t$ tokens, correspondingly. 

\paragraph{Step 1: Tokenization} The comprehensiveness of the pre-training corpus is important to obtain a well-trained token representation. 
An unbalanced corpus makes it hard to learn the representation of tokens in the tail of vocabulary. 
Thus, the corpus used in this work is empirically composed of multilingual corpus ``CulturaX'' [40\%] \citep{nguyen-etal-2024-culturax}, code corpus ``The Stack'' [30\%] \citep{kocetkov2023stack}, and math corpus ``Proof-Pile-2'' [30\%] \citep{azerbayev2024llemma}. 
We tokenize the mixed corpus using various tokenizers and obtain multiple sequences of token IDs for the same corpus. 
The default amount of tokens used in this step is 1B, which is investigated in Appendix \ref{appendix:glove}. 

\paragraph{Step 2: Token Representation Learning} We adopt GloVe \citep{pennington-etal-2014-glove} to learn the representation of tokens from the first step. 
The main reason is that GloVe considers more global statistical information than those slide window methods like CBOW and FastText \citep{mikolov2013cbow, mikolov2013skip, bojanowski2017fasttext}. 
The details of training settings for GloVe vectors refer to Appendix \ref{appendix:param}. 

\paragraph{Step 3: Token Alignment} Based on the assumption that token representations capture the semantic information in the token, we align token IDs using the pair-wise cosine similarity of learned token representations. 
It should be noted that the IDs of tokens belonging to both vocabularies are directly replaced without the need to align. 
$M_{s\to t}$ denotes the learned token-token alignment matrix, which records the pair-wise similarity of each source token and target token. 
It can serve as the one-to-one mapping function for each source/target token to find the most similar token from the target/source vocabulary. 

\begin{figure*}[t]
\centering
\includegraphics [width=0.9\textwidth]{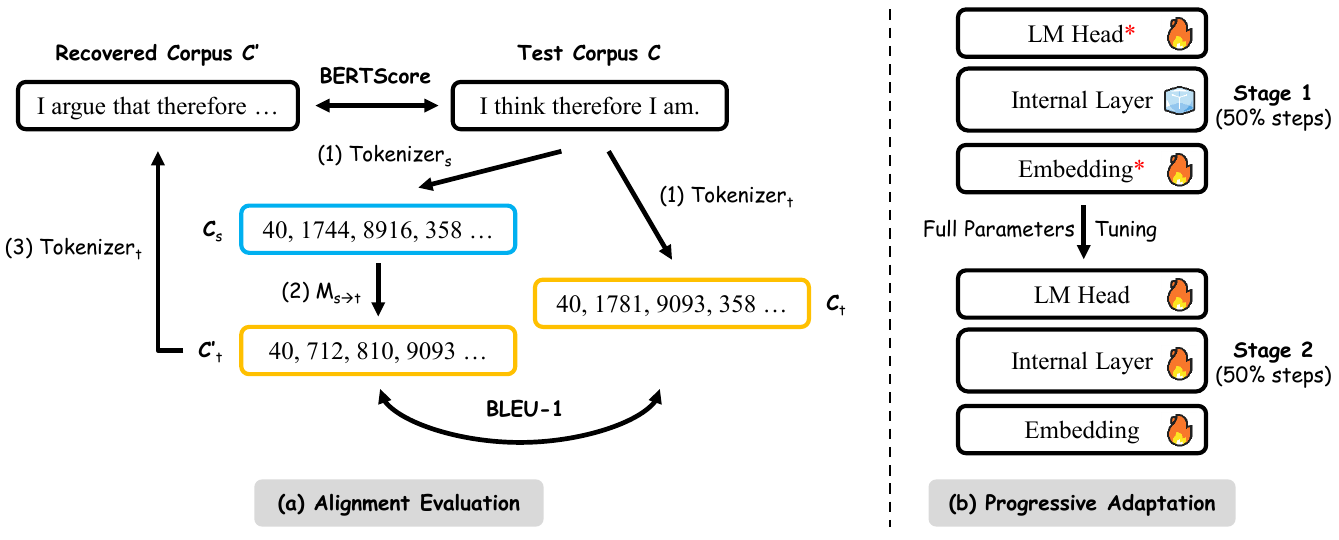}
\vspace{-3mm}
\caption{\label{fig:eval_tune} (a) We choose BLEU-1 and BERTScore to evaluate the performance of alignment matrix $M_{s\to t}$ (b) Embedding and lm$\_$head are tuned at the first half part of the process, followed by full parameter tuning. {\color{red} *} indicates the parameter of each target token is first initialized from the most similar source token by alignment matrix $M_{s\to t}$.}
\vspace{-4mm}
\end{figure*}

\subsection{Alignment Evaluation}
\label{sec:align_eval}

\hyperref[fig:eval_tune]{Figure 3(a)} illustrates our metrics to evaluate the performance of alignment matrix $M_{s\to t}$. 
We first tokenize the test corpus $\mathcal{C}$ using different tokenizers, which results in $\mathcal{C}_s$ and $\mathcal{C}_t$. 
The token ID corpus $\mathcal{C}_s$ from the source tokenizer is converted to its most similar target token ID by alignment matrix $M_{s\to t}$, and comes to the corpus $\mathcal{C}_{t}^{'}$. 
From the view of token ID matching, the higher BLEU-1 score between $\mathcal{C}_{t}^{'}$ and the corpus $\mathcal{C}_t$ from the Tokenizer${}_{t}$, the better alignment matrix $M_{s\to t}$ is. 

We further propose a semantic evaluation metric: 
It de-tokenizes the target token ID corpus $\mathcal{C}_{t}^{'}$ using Tokenizer${}_{t}$ into the recovered text corpus $\mathcal{C}^{'}$, and evaluates the semantic similarity between $\mathcal{C}^{'}$ and original corpus $\mathcal{C}$ using BERTScore. 
The better alignment matrix $M_{s\to t}$ learned preserves more semantics in the test corpus $\mathcal{C}$, bringing higher BERTScore of the recovered $\mathcal{C}^{'}$ and $\mathcal{C}$. 

\subsection{Progressive Adaptation}
Given the alignment matrix $M_{s\to t}$, the parameters of each token in the target vocabulary are initialized from the ones of the most similar source token. 
We find that these re-arranged embeddings and lm$\_$head provide a good initialization for the new model (Section \ref{sec:cross_transfer}). 
\hyperref[fig:eval_tune]{Figure 3(b)} illustrates the two-stage tuning for an LLM to adapt to the new vocabulary. 
The re-arranged embedding and lm$\_$head are tuned first to avoid loss spike and improve the training stability (Figure \ref{fig:loss_2stage}). 
The other parameters of internal layers are further tuned together in the last half-part process. 


\section{Experiments}
\subsection{Experiments Settings}

\paragraph{Large Language Models} We adopt the fully open-source language model series Pythia \citep{biderman2023pythia} as base models in this work. 
It is noted that we do not intend to achieve state-of-the-art large language model performance but rather investigate an efficient method to replace the English-centric tokenizer like Pythia. 
To transfer token-level knowledge from other capable large language models, tokenizers and vocabularies of Gemma \citep{team2024gemma}, Qwen2 \citep{yang2024qwen2}, LLaMA2 \citep{touvron2023llama2}, and LLaMA3 \citep{meta2024llama3} are selected as the target to replace. 
We report hyper-parameters in Appendix \ref{appendix:param}. 

\paragraph{Corpus} To reduce the risk of distribution shift from the training data, we choose the vanilla pre-training corpus Pile \citep{gao2020pile} of Pythia in the fine-tuning process. 
We also investigate the robustness of the corpus used in the vocabulary alignment by replacing it with Slimpajama \citep{cerebras2023slimpajama}. 
Corpora of downstream tasks and multiple languages are applied in cross-lingual and cross-model knowledge transfer experiments (Section \ref{sec:cross_transfer} and \ref{sec:tok_dist}). 

\paragraph{Evaluation Tasks} Following the common practices to evaluate large language models \citep{lin-etal-2022-shot, biderman2023pythia, zhang2024tinyllama}, there are 10 datasets, including commonsense reasoning \citep{clark2018arc, mihaylov-etal-2018-suit, zellers-etal-2019-hellaswag, ponti-etal-2020-xcopa, bisk-etal-2020-piqa, Sakaguchi-et-al-2020-winogrande} and reading comprehension \citep{clark-etal-2019-boolq} tasks, used in this work. 
To avoid the randomness from the prompt and evaluation method, we adopt the default prompt from the commonly used language model evaluation harness framework \citep{eval-harness}. 
Further information about the evaluation tasks is reported in Appendix \ref{app:eval_tasks}. 

\begin{table*}[htp]

\renewcommand\arraystretch{0.7}

\centering
\scriptsize

\setlength{\tabcolsep}{1.4mm}

 \begin{tabu}{l|ccccc|ccccc|ccc|c}
 
 \toprule[1.2pt]
  
  \multicolumn{1}{c}{ } & \multicolumn{5}{c}{\textbf{High}} & \multicolumn{5}{c}{\textbf{Medium}} & \multicolumn{3}{c}{\textbf{Low}} & \multicolumn{1}{c}{ } \\

  \cmidrule(r){2-6} \cmidrule(r){7-11} \cmidrule(r){12-14} \noalign{\smallskip}

 \multicolumn{1}{c}{\textbf{Model}} & \textbf{ar} & \textbf{de} & \textbf{en} & \textbf{ja} & \textbf{zh} & \textbf{bn} & \textbf{ko} & \textbf{th} & \textbf{uk} & \textbf{vi} & \textbf{ta} & \textbf{te} & \multicolumn{1}{c}{\textbf{ur}} & \multicolumn{1}{c}{\textbf{Avg} $\downarrow$} \\

\midrule[0.8pt]

$\text{Qwen2}_{\text{1.5B}}$
&$4.7$&$11.1$&$15.7$&$6.0$&$4.6$&$2.4$&$3.3$&$2.6$&$5.7$&$3.3$&$2.8$&$3.4$&$4.0$&$5.3$\\

\midrule[0.8pt]

$\text{Pythia}_{\text{1B}}$
&$7.6$&$15.4$&$\textbf{21.7}$&$9.9$&$13.2$&$3.4$&$5.6$&$4.3$&$6.7$&$6.3$&$2.9$&$3.3$&$5.8$&$8.2$\\

   \cdashlinelr{1-15}

$\text{\ w/\ }{\text{Focus Init.}}$
&$4.1e^3$&$1.7e^5$&$1.8e^6$&$2.1e^4$&$9.6e^2$&$6.5e^4$&$1.0e^3$&$5.6e^3$&$1.6e^6$&$8.4e^2$&$5.0e^4$&$1.9e^5$&$1.9e^5$&$3.1e^5$\\ 

$\text{\ \ \ \ +\ LAT}$&$8.3$&$27.1$&$59.7$&$14.0$&$14.0$&$3.6$&$5.9$&$3.8$&$7.3$&$5.9$&$3.5$&$3.6$&$4.3$&$12.4$\\

$\text{\ w/\ }{\text{ZeTT Init.}}$
&$3.0e^2$&$4.2e^2$&$1.3e^2$&$1.2e^3$&$2.4e^2$&$3.0e^2$&$2.4e^2$&$3.3e^2$&$2.5e^2$&$2.0e^2$&$2.4e^2$&$1.8e^2$&$4.7e^2$&$3.4e^2$\\

$\text{\ \ \ \ +\ LAT}$&$7.1$&$15.7$&$26.4$&$10.0$&$10.3$&$2.8$&$5.0$&$3.6$&$5.9$&$4.9$&$2.6$&$2.7$&$4.2$&$7.8$\\


$\text{\ w/\ }{\text{TokAlign Init.}}$
&$1.2e^2$&$2.2e^2$&$1.0e^2$&$3.6e^2$&$1.2e^2$&$46.5$&$60.1$&$70.8$&$1.5e^2$&$49.2$&$61.0$&$1.1e^2$&$50.9$&$1.2e^2$\\

$\text{\ \ \ \ +\ LAT}$&$\textbf{6.3}$&$\textbf{13.9}$&$23.6$&$\textbf{8.9}$&$\textbf{9.0}$&$\textbf{2.4}$&$\textbf{4.4}$&$\textbf{3.2}$&$\textbf{5.2}$&$\textbf{4.4}$&$\textbf{2.3}$&$\textbf{2.4}$&$\textbf{3.7}$&$\textbf{6.9}$\\

\midrule[0.8pt]

$\text{Qwen2}_{\text{7B}}$&$3.9$&$8.1$&$11.8$&$4.9$&$3.8$&$2.1$&$2.9$&$2.3$&$3.8$&$2.9$&$2.3$&$2.6$&$3.3$&$4.2$\\

\midrule[0.8pt]

 $\text{Pythia}_{\text{6.9B}}$
&$5.9$&$10.8$&$\textbf{16.7}$&$7.9$&$9.9$&$3.0$&$4.6$&$3.7$&$4.9$&$4.9$&$2.6$&$2.9$&$4.8$&$6.3$\\

   \cdashlinelr{1-15}

$\text{\ w/\ }{\text{Focus Init.}}$
&$6.9e^3$&$1.6e^5$&$1.2e^6$&$2.4e^4$&$1.3e^3$&$2.5e^4$&$7.2e^2$&$3.3e^3$&$1.9e^6$&$7.9e^2$&$1.7e^4$&$1.5e^5$&$1.2e^5$&$2.8e^5$\\

$\text{\ \ \ \ +\ LAT}$&$6.8$&$17.6$&$39.3$&$10.8$&$11.1$&$2.5$&$5.0$&$3.3$&$5.2$&$4.8$&$2.3$&$2.5$&$3.7$&$8.8$\\


$\text{\ w/\ }{\text{TokAlign Init.}}$
&$1.2e^2$&$1.9e^2$&$81.4$&$3.7e^2$&$1.3e^2$&$52.5$&$53.3$&$66.2$&$1.4e^2$&$49.2$&$46.4$&$92.1$&$48.7$&$1.1e^2$\\

$\text{\ \ \ \ +\ LAT}$&$\textbf{5.2}$&$\textbf{9.9}$&$17.8$&$\textbf{7.4}$&$\textbf{7.9}$&$\textbf{2.1}$&$\textbf{3.8}$&$\textbf{2.8}$&$\textbf{4.0}$&$\textbf{3.7}$&$\textbf{2.1}$&$\textbf{2.1}$&$\textbf{3.1}$&$\textbf{5.5}$\\

\midrule[0.8pt]

$\Delta\textbf{\ Length (\%) $\downarrow$}$&$-44.5$&$-13.1$&$-0.8$&$-32.4$&$-50.0$&$-22.2$&$-52.2$&$-46.1$&$-15.5$&$-51.7$&$-20.3$&$-2.9$&$-28.5$&$-29.2$\\

\bottomrule[1.2pt]
\end{tabu}
\vspace{-2mm}
\caption{\label{tab:multilingual_ppl} The normalized perplexity on the valid corpus of CulturaX. The perplexity is normalized to the vocabulary of Pythia following \citet{wei2023skywork}. 
``\textbf{High}'', ``\textbf{Medium}'', and ``\textbf{Low}'' indicates the available amount of linguistic resources. ``w/ xxx Init.'' denotes the performance of the model after initialization without any tuning steps. 
}
\vspace{-2mm}
\end{table*}

\paragraph{Baselines} We introduce the following vocabulary adaptation methods as baseline methods in this work:
\begin{itemize}
    \item \textbf{Random Initialization} for each token $t \in \{\mathcal{V}_t\setminus (\mathcal{V}_t \cap \mathcal{V}_s)\}$ employs the default initialization method of huggingface Transformers and reuses the parameters of token $t \in \{\mathcal{V}_t\cap \mathcal{V}_s\}$, which belongs to both vocabularies.
    \item \textbf{Random Permutation} initializes each token $t \in \{\mathcal{V}_t\setminus (\mathcal{V}_t \cap \mathcal{V}_s)\}$ using the parameter of randomly chosen token from the source vocabulary. The parameters of shared tokens are also reused. 
    \item \textbf{Multivariate} initializes each token $t \in \{\mathcal{V}_t\setminus (\mathcal{V}_t \cap \mathcal{V}_s)\}$ by sampling from the multivariate Gaussian distribution with the mean and covariance of source embedding $\textit{E}_s$. 
    \item \textbf{Mean} use the mean of source embedding $\textit{E}_s$ to initialize all tokens $t \in \{\mathcal{V}_t\setminus (\mathcal{V}_t \cap \mathcal{V}_s)\}$. 
    \item \textbf{WECHSEL} \citep{minixhofer-etal-2022-wechsel} linearly transfers embeddings of source tokens into target tokens by tokenizing and recomposing additional word embeddings $\textit{W}^{s}$ and $\textit{W}^{t}$, which are aligned with a bilingual dictionary. 
    \item \textbf{OFA} \citep{liu-etal-2024-ofa} factorizes the embeddings of source model $\textit{E}_s$ into the primitive embedding $\textit{P}$ and source coordinate $\textit{F}_s$ that is further re-composed by multilingual word embedding $\textit{W}$ to the target coordinate $\textit{F}_t$. The assembled primitive embedding $\textit{P}$ and target coordinate $\textit{F}_t$ yield the target embedding $\textit{E}_t$. 
    \item \textbf{Focus} \citep{dobler-de-melo-2023-focus} initializes the embedding parameters of token $t \in \{\mathcal{V}_t\setminus (\mathcal{V}_t \cap \mathcal{V}_s)\}$ using the weighted sum of the ones from the token $t \in \{\mathcal{V}_t\cap \mathcal{V}_s\}$. It largely depends on the size of $\| \mathcal{V}_t\cap \mathcal{V}_s \|$, and performs poorly when the overlapping percentage of $\mathcal{V}_t$ and $\mathcal{V}_s$ is low. 
    \item \textbf{ZeTT} \citep{minixhofer2024zett} trains an additional hypernetwork $\textit{H}_{\theta}$ to generate the parameters for each token $t \in \mathcal{V}_t$. The added hypernetwork brings a lot of training costs. 
\end{itemize}

\begin{table*}[thp]

\renewcommand\arraystretch{0.9}

\centering
\scriptsize

\setlength{\tabcolsep}{0.9mm}

 \begin{tabu}{l|ccccccc|ccccc|ccc|cccc|c}
 
 \toprule[1.2pt]
  
  \multicolumn{1}{c}{ } & \multicolumn{7}{c}{\textbf{XNLI}} & \multicolumn{5}{c}{\textbf{PAWS-X}} & \multicolumn{3}{c}{\textbf{XCOPA}}  & \multicolumn{4}{c}{\textbf{XStoryCloze}}  & \multicolumn{1}{c}{} \\

  \cmidrule(r){2-8} \cmidrule(r){9-13} \cmidrule(r){14-16} \cmidrule(r){17-20} \noalign{\smallskip}

 \multicolumn{1}{c}{\textbf{Model}} & \textbf{en} & \textbf{de} & \textbf{zh} & \textbf{ar} & \textbf{th} & \textbf{vi} & \textbf{ur} & \textbf{de} & \textbf{en} & \textbf{ja} & \textbf{ko} & \textbf{zh} & \textbf{th} & \textbf{vi} & \textbf{ta} & \textbf{en} & \textbf{zh} & \textbf{ar} & \multicolumn{1}{c}{\textbf{te}} & \multicolumn{1}{c}{\textbf{Avg}} \\

\midrule[0.8pt]

$\text{Pythia}_{\text{1B}}$&$\textbf{51.0}$&$37.8$&$42.6$&$35.9$&$34.8$&$37.0$&$34.7$&$49.6$&$49.3$&$54.8$&$\textbf{54.9}$&$52.9$&$54.0$&$53.2$&$55.4$&$\textbf{64.3}$&$48.6$&$48.0$&$52.9$&$48.0$\\

   \cdashlinelr{1-21}

$\text{\ \ \ w/\ }{\text{Focus Init.}}$&$32.8$&$32.2$&$33.6$&$33.6$&$33.5$&$32.0$&$32.8$&$44.8$&$44.9$&$45.7$&$44.8$&$44.7$&$52.4$&$48.6$&$\textbf{57.0}$&$45.9$&$47.8$&$\textbf{48.8}$&$46.5$&$42.2$\\

$\text{\ \ \ \ +\ LAT}$&$46.0$&$35.1$&$34.9$&$32.9$&$32.5$&$35.4$&$34.7$&$50.6$&$45.5$&$55.9$&$53.4$&$\textbf{55.3}$&$53.8$&$52.6$&$55.4$&$55.8$&$48.8$&$47.6$&$50.4$&$46.1$\\

$\text{\ \ \ w/\ }{\text{ZeTT Init.}}$&$45.9$&$34.6$&$32.9$&$32.8$&$33.5$&$33.6$&$34.5$&$51.5$&$50.3$&$54.8$&$51.5$&$53.5$&$52.6$&$48.2$&$55.6$&$53.2$&$46.9$&$46.9$&$48.1$&$45.3$\\

$\text{\ \ \ \ +\ LAT}$&$48.6$&$38.6$&$40.6$&$36.9$&$36.0$&$39.3$&$35.1$&$53.0$&$51.0$&$55.8$&$53.8$&$\textbf{55.3}$&$\textbf{55.8}$&$50.8$&$54.0$&$60.3$&$49.3$&$47.2$&$52.1$&$48.1$\\

$\text{\ \ \ w/\ }{\text{TokAlign Init.}}$
&$49.9$&$36.6$&$33.2$&$31.8$&$33.2$&$34.4$&$34.4$&$52.4$&$\textbf{52.1}$&$\textbf{56.1}$&$54.7$&$\textbf{55.3}$&$53.6$&$48.0$&$55.2$&$61.0$&$47.6$&$47.1$&$51.0$&$46.7$\\

$\text{\ \ \ \ +\ LAT}$&$50.9$&$\textbf{39.3}$&$\textbf{42.7}$&$\textbf{37.4}$&$\textbf{37.4}$&$\textbf{40.3}$&$\textbf{35.7}$&$\textbf{54.6}$&$50.2$&$55.9$&$\textbf{54.9}$&$\textbf{55.3}$&$55.2$&$\textbf{53.6}$&$53.6$&$64.0$&$\textbf{51.1}$&$47.8$&$\textbf{53.5}$&$\textbf{49.1}$\\

\midrule[0.8pt]

$\text{Pythia}_{\text{6.9B}}$&$54.4$&$\textbf{39.0}$&$\textbf{46.2}$&$39.3$&$39.8$&$39.3$&$36.4$&$43.8$&$40.2$&$50.2$&$54.2$&$50.2$&$\textbf{56.2}$&$54.4$&$52.2$&$\textbf{70.4}$&$53.9$&$\textbf{50.3}$&$53.8$&$48.6$\\

\cdashlinelr{1-21}

$\text{\ \ \ w/\ }{\text{Focus Init.}}$&$31.5$&$31.3$&$33.0$&$32.6$&$33.4$&$32.2$&$32.6$&$44.8$&$42.4$&$52.7$&$45.5$&$44.7$&$52.2$&$48.6$&$\textbf{55.6}$&$44.5$&$47.1$&$47.8$&$47.1$&$42.1$\\

$\text{\ \ \ \ +\ LAT}$&$52.6$&$34.9$&$36.6$&$35.1$&$33.6$&$39.0$&$34.5$&$\textbf{51.1}$&$43.8$&$55.9$&$55.3$&$55.4$&$54.2$&$52.4$&$53.8$&$61.0$&$48.7$&$47.7$&$53.7$&$47.3$\\

$\text{\ \ \ w/\ }{\text{TokAlign Init.}}$&$53.3$&$36.3$&$35.0$&$34.6$&$34.6$&$33.0$&$33.8$&$48.8$&$44.6$&$\textbf{56.2}$&$55.7$&$55.3$&$54.6$&$52.2$&$54.6$&$66.8$&$48.6$&$47.7$&$50.0$&$47.1$\\

$\text{\ \ \ \ +\ LAT}$&$\textbf{55.2}$&$35.8$&$43.5$&$\textbf{40.4}$&$\textbf{40.2}$&$\textbf{43.0}$&$\textbf{37.1}$&$43.2$&$\textbf{45.8}$&$55.8$&$\textbf{55.8}$&$\textbf{55.5}$&$54.6$&$\textbf{57.0}$&$54.6$&$70.2$&$\textbf{54.4}$&$49.3$&$\textbf{53.9}$&$\textbf{49.7}$\\

\bottomrule[1.2pt]
\end{tabu}
\vspace{-2mm}

\caption{\label{tab:multilingual_0_shot} Zero-shot in-context learning results of cross-lingual transfer. 
Refer to Table \ref{tab:multilingual_5_shot} for few-shot results. 
}

\vspace{-2mm}

\end{table*}

\begin{table*}[thp]

\renewcommand\arraystretch{0.9}

\centering
\scriptsize

\setlength{\tabcolsep}{1.6mm}

 \begin{tabu}{l|cc|cc|cc|cc|cc|cc|cc}
 
 \toprule[1.2pt]
  
  \multicolumn{1}{c}{ } & \multicolumn{2}{c}{\textbf{ARC-E}} & \multicolumn{2}{c}{\textbf{BoolQ}} & \multicolumn{2}{c}{\textbf{HellaSwag}} & \multicolumn{2}{c}{\textbf{OpenbookQA}} & \multicolumn{2}{c}{\textbf{PIQA}}  & \multicolumn{2}{c}{\textbf{WinoGrande}} & \multicolumn{2}{c}{\textbf{Avg}} \\

  \cmidrule(r){2-3} \cmidrule(r){4-5} \cmidrule(r){6-7} \cmidrule(r){8-9} \cmidrule(r){10-11} \cmidrule(r){12-13} \cmidrule(r){14-15} \noalign{\smallskip}

 \multicolumn{1}{c}{\textbf{Model}} & \textbf{0} & \textbf{5} & \textbf{0} & \textbf{5} & \textbf{0} & \textbf{5} & \textbf{0} & \textbf{5} & \textbf{0} & \textbf{5} & \textbf{0} & \textbf{5} & \textbf{0} &  \multicolumn{1}{c}{\textbf{5}} \\

\midrule[0.8pt]

$\text{Pythia}_{\text{1B}}$
&$56.82$&$58.71$&$60.43$&$57.37$&$37.68$&$37.66$&$18.80$&$19.00$&$70.40$&$71.49$&$53.20$&$52.01$&$49.55$&$49.37$\\

$\text{\ \ \ +\ Direct tuning}$
&$57.49$&$55.64$&$70.70$&$72.11$&$41.24$&$41.60$&$25.40$&$28.40$&$69.04$&$70.08$&$54.70$&$54.78$&$53.10$&$53.77$\\

$\text{\ \ \ +\ Sentence distill}$
&$52.27$&$53.41$&$67.49$&$67.06$&$39.03$&$39.08$&$21.80$&$22.80$&$66.97$&$68.99$&$51.85$&$52.17$&$49.90$&$50.58$\\

   \cdashlinelr{1-15}
   
$\text{\ \ \ w/\ }{\text{Gemma}_{\text{7B}}}$
&$55.39$&$56.99$&$67.19$&$69.69$&$36.53$&$37.26$&$19.00$&$22.80$&$68.82$&$69.21$&$52.33$&$53.51$&$49.88$&$51.58$\\

$\text{\ \ \ w/\ }{\text{Qwen2}_{\text{7B}}}$
&$62.33$&$63.17$&$70.18$&$72.54$&$41.58$&$42.21$&$22.00$&$\textbf{28.20}$&$\textbf{73.01}$&$73.18$&$55.01$&$55.56$&$54.02$&$55.81$\\

$\text{\ \ \ w/\ }{\text{LLaMA3}_{\text{8B}}}$
&$\textbf{64.02}$&$\textbf{64.56}$&$\textbf{73.91}$&$\textbf{74.19}$&$\textbf{42.11}$&$\textbf{42.34}$&$\textbf{24.20}$&$27.60$&$72.74$&$\textbf{73.83}$&$\textbf{55.49}$&$\textbf{56.43}$&$\textbf{55.41}$&$\textbf{56.49}$\\

\midrule[0.8pt]

 $\text{Pythia}_{\text{6.9B}}$
&$65.99$&$69.23$&$62.84$&$62.02$&$47.56$&$47.64$&$25.00$&$27.00$&$74.65$&$75.41$&$60.46$&$62.43$&$56.08$&$57.29$\\

$\text{\ \ \ +\ Direct tuning}$
&$66.25$&$66.20$&$79.30$&$78.87$&$52.21$&$53.39$&$33.20$&$33.00$&$72.91$&$74.48$&$62.90$&$61.72$&$61.13$&$61.28$\\

$\text{\ \ \ +\ Sentence distill}$
&$61.70$&$65.36$&$76.64$&$76.88$&$48.98$&$51.33$&$28.20$&$30.40$&$70.18$&$71.55$&$58.96$&$62.19$&$57.44$&$59.62$\\

   \cdashlinelr{1-15}

$\text{\ \ \ w/\ }{\text{Gemma}_{\text{7B}}}$
&$67.59$&$68.94$&$76.06$&$75.66$&$47.83$&$48.36$&$28.40$&$31.40$&$73.78$&$75.52$&$59.04$&$64.17$&$58.78$&$60.67$\\

$\text{\ \ \ w/\ }{\text{Qwen2}_{\text{7B}}}$
&$\textbf{71.72}$&$\textbf{73.27}$&$\textbf{79.85}$&$\textbf{80.00}$&$\textbf{50.78}$&$\textbf{51.12}$&$\textbf{29.20}$&$\textbf{34.00}$&$\textbf{77.26}$&$\textbf{77.91}$&$\textbf{61.33}$&$\textbf{64.56}$&$\textbf{61.69}$&$\textbf{63.48}$\\

$\text{\ \ \ w/\ }{\text{LLaMA3}_{\text{8B}}}$
&$67.05$&$69.78$&$77.83$&$78.78$&$48.83$&$50.15$&$26.00$&$32.00$&$74.21$&$76.22$&$60.22$&$60.93$&$59.02$&$61.31$\\

\bottomrule[1.2pt]
\end{tabu}
\vspace{-2mm}
\caption{\label{tab:token_distill_res} The main results of token-level distillation on six downstream tasks with only 235M tokens. ``+Sentence distill'' denotes the sentence-level distillation results with Qwen2${}_{\text{7B}}$\citep{yang2024qwen2}, which fine-tunes on the output from Qwen2${}_{\text{7B}}$ given questions as prompt.
}
\vspace{-2mm}
\end{table*}



\subsection{Main Results}
\label{sec:main_res}
We first report the final results of two applications after replacing vocabulary: cross-lingual transfer (Section \ref{sec:cross_transfer}) and cross-model knowledge transfer (Section \ref{sec:tok_dist}), then show vocabulary adaptation results of methods (Section \ref{sec:vocab_adapt}). 

\subsubsection{Cross-lingual Transfer}
\label{sec:cross_transfer}
When applied to new domains or languages, tokenizers with higher compression rates can speed up the learning and inference of large language models. 
From the view of token co-occurrence, tokens from other languages can be aligned and initialized by the tokens with similar semantics in the source vocabulary, which can boost the cross-lingual knowledge transfer. 
Therefore, we replace the English-centric tokenizer of Pythia with the one of Qwen2 to evaluate the performance on cross-lingual transfer settings.

As shown in Table \ref{tab:multilingual_ppl}, the perplexity of Pythia initialized using TokAlign ($1.2e^2$) is significantly better than other two strong baseline methods Focus ($2.9e^5$) and ZeTT($3.4e^2$). 
The length of tokens after text tokenization has reduced by 29.2\% on average across these languages. 
After only 2k steps of \textbf{L}anguage \textbf{A}daptation \textbf{T}uning (``+LAT''), TokAlign improved 14.5\% over the vanilla model on average, while Focus still performed worse.
It is noted that the performance of Pythia using TokAlign on three low-resource languages even outperforms the ones of Qwen2 with a similar parameter amount. 


Table \ref{tab:multilingual_0_shot} and \ref{tab:multilingual_5_shot} in Appendix \ref{appendix:cross_lingual_transfer} further report zero-shot and few-shot in-context learning results on four multilingual datasets. 
We can find that TokAlign brings a better-initialized model than the baseline method Focus (+4.4\%), and transfers the knowledge into other languages like Japanese (ja, +2.3\%) and Vietnamese (vi, +2.2\%). 

It is interesting to find that the perplexity of Pythia${}_{\text{1B}}$ initialized by TokAlign reaches 1.2${e}^{2}$, while the in-context learning results are comparable with the ones of Focus after adapting on the multilingual corpus. 
We argue that it arises from the reserved English ability with TokAlign (54.2\%), which significantly outperforms Focus (40.8\%). 

\begin{table*}[thp]

\renewcommand\arraystretch{0.9}

\centering
\scriptsize

\setlength{\tabcolsep}{0.7mm}

 \begin{tabu}{lr|cc|cc|cc|cc|cc|cc|cc}
 
 \toprule[1.2pt]
  
  \multicolumn{2}{c}{ } & \multicolumn{2}{c}{\textbf{ARC-E}} & \multicolumn{2}{c}{\textbf{BoolQ}} & \multicolumn{2}{c}{\textbf{HellaSwag}} & \multicolumn{2}{c}{\textbf{OpenbookQA}} & \multicolumn{2}{c}{\textbf{PIQA}}  & \multicolumn{2}{c}{\textbf{WinoGrande}} & \multicolumn{2}{c}{\textbf{Avg}} \\

  \cmidrule(r){3-4} \cmidrule(r){5-6} \cmidrule(r){7-8} \cmidrule(r){9-10} \cmidrule(r){11-12} \cmidrule(r){13-14} \cmidrule(r){15-16} \noalign{\smallskip}

 \multicolumn{1}{c}{\textbf{Model}} & \multicolumn{1}{c}{\tiny $\#$\textbf{GPU Hour}} & \textbf{0} & \textbf{5} & \textbf{0} & \textbf{5} & \textbf{0} & \textbf{5} & \textbf{0} & \textbf{5} & \textbf{0} & \textbf{5} & \textbf{0} & \textbf{5} & \textbf{0} &  \multicolumn{1}{c}{\textbf{5}} \\

\midrule[0.8pt]

$\text{Pythia}_{\text{1B}}$
&\multicolumn{1}{c}{$-$}&$56.82$&$58.71$&$60.43$&$57.37$&$37.68$&$37.66$&$18.80$&$19.00$&$70.40$&$71.49$&$53.20$&$52.01$&$49.55$&$49.37$\\

   \cdashlinelr{1-16}

$\text{\ w$/$\ Rand. Init.}$
&$99.70\ $&$31.36$&$31.61$&$37.83$&$49.11$&$26.35$&$26.40$&$14.00$&$12.60$&$54.57$&$55.33$&$49.17$&$49.17$&$35.55$&$37.37$\\

$\text{\ w$/$\ Rand. Perm.}$
&$99.70\ $&$31.69$&$32.95$&$37.77$&$54.80$&$26.43$&$26.39$&$14.00$&$12.60$&$55.50$&$55.98$&$47.04$&$50.67$&$35.40$&$38.90$\\

$\text{\ w$/$\ Multivariate}$
&$99.70\ $&$32.79$&$34.18$&$45.08$&$49.72$&$27.67$&$27.87$&$15.20$&$16.20$&$56.09$&$57.83$&$50.51$&$50.12$&$37.89$&$39.32$\\

$\text{\ w$/$\ Mean}$
&$99.70\ $&$44.87$&$46.97$&$53.39$&$55.20$&$31.59$&$31.67$&$16.20$&$17.00$&$61.32$&$62.46$&$49.25$&$51.85$&$42.77$&$44.19$\\

$\text{\ w$/$\ OFA}$
&$99.70\ $&$38.17$&$37.79$&$55.14$&$52.35$&$28.29$&$28.62$&$14.40$&$12.20$&$58.43$&$58.54$&$49.96$&$50.99$&$40.73$&$40.08$\\

$\text{\ w$/$\ WECHSEL}$
&$99.70\ $&$43.35$&$45.33$&$56.61$&$54.34$&$32.53$&$32.41$&$14.80$&$16.20$&$61.70$&$62.89$&$52.01$&$52.72$&$43.50$&$43.98$\\

$\text{\ w$/$\ Focus}$
&$99.70\ $&$46.55$&$48.95$&$56.21$&$\textbf{55.78}$&$32.27$&$32.46$&$19.20$&$18.00$&$63.82$&$64.80$&$51.70$&$51.78$&$44.96$&$45.29$\\

$\text{\ w$/$\ ZeTT}$
&$418.94\ $&$47.14$&$49.03$&$57.06$&$53.70$&$34.06$&$34.06$&$18.40$&$19.40$&$64.15$&$65.34$&$52.09$&$51.22$&$45.48$&$45.46$\\

$\text{\ w$/$\ TokAlign}$
&$99.70\ $&$\textbf{54.46}^{*}$&$\textbf{56.86}^{*}$&$58.90^{*}$&$52.26$&$36.16^{*}$&$36.27^{*}$&$\textbf{21.00}^{*}$&$\textbf{20.20}^{*}$&$\textbf{67.74}^{*}$&$\textbf{68.50}^{*}$&$52.25^{*}$&$50.91$&$48.42$&$47.50$\\

$\text{\ \ \ \ \ \ w$/$\ SlimPajama}$
&$99.70\ $&$53.54$&$55.68$&$57.55$&$53.85$&$36.10$&$35.99$&$19.40$&$\textbf{20.20}$&$67.03$&$67.52$&$52.09$&$51.22$&$47.62$&$47.41$\\

$\text{\ \ \ \ \ \ +\ Align Rep.}$
&$99.70\ $&$54.25$&$56.65$&$\textbf{59.33}$&$54.68$&$\textbf{37.08}$&$\textbf{36.91}$&$20.20$&$19.40$&$67.36$&$68.17$&$\textbf{54.38}$&$\textbf{52.80}$&$\textbf{48.77}$&$\textbf{48.10}$\\

\midrule[0.8pt]

 $\text{Pythia}_{\text{2.8B}}$
&\multicolumn{1}{c}{$-$}&$63.80$&$67.00$&$63.91$&$65.14$&$45.32$&$45.04$&$24.00$&$25.20$&$74.05$&$74.43$&$58.64$&$60.77$&$54.95$&$56.26$\\

   \cdashlinelr{1-16}

$\text{\ w$/$\ Rand. Init.}$
&$194.78\ $&$30.47$&$32.91$&$38.20$&$51.07$&$26.46$&$26.69$&$14.40$&$13.20$&$55.17$&$55.06$&$48.30$&$50.51$&$35.50$&$38.24$\\

$\text{\ w$/$\ Rand. Perm.}$
&$194.78\ $&$31.48$&$31.86$&$37.83$&$50.46$&$26.48$&$26.49$&$13.60$&$14.40$&$54.03$&$54.95$&$50.20$&$48.86$&$35.60$&$37.84$\\

$\text{\ w$/$\ OFA}$
&$194.78\ $&$50.13$&$54.12$&$60.89$&$61.47$&$36.39$&$36.88$&$18.00$&$19.00$&$65.18$&$64.80$&$54.06$&$54.85$&$47.44$&$48.52$\\

$\text{\ w$/$\ WECHSEL}$
&$194.78\ $&$52.48$&$54.92$&$59.42$&$56.76$&$36.79$&$37.30$&$19.20$&$20.80$&$64.04$&$64.25$&$56.43$&$55.72$&$48.06$&$48.29$\\

$\text{\ w$/$\ Focus}$
&$194.78\ $&$54.29$&$58.16$&$61.44$&$62.84$&$38.38$&$39.09$&$20.00$&$20.20$&$68.44$&$68.28$&$54.62$&$56.04$&$49.53$&$50.77$\\

$\text{\ w$/$\ ZeTT}$
&$855.96\ $&$57.15$&$59.42$&$61.68$&$62.05$&$42.17$&$42.25$&$21.80$&$23.60$&$71.11$&$71.16$&$56.59$&$59.19$&$51.75$&$52.95$\\

$\text{\ w$/$\ TokAlign}$
&$194.78\ $&$61.62^{*}$&$65.15^{*}$&$63.82^{*}$&$65.47^{*}$&$43.13^{*}$&$43.18^{*}$&$\textbf{23.40}^{*}$&$\textbf{25.80}^{*}$&$72.14^{*}$&$72.42^{*}$&$\textbf{58.17}^{*}$&$\textbf{61.17}^{*}$&$53.71$&$55.53$\\

$\text{\ \ \ \ \ \ +\ Align Rep.}$
&$194.78\ $&$\textbf{61.66}$&$\textbf{65.66}$&$\textbf{64.56}$&$\textbf{65.66}$&$\textbf{43.97}$&$\textbf{44.09}$&$22.40$&$25.00$&$\textbf{73.01}$&$\textbf{73.23}$&$58.09$&$60.54$&$\textbf{53.95}$&$\textbf{55.70}$\\

\bottomrule[1.2pt]
\end{tabu}
\vspace{-1mm}
\caption{\label{tab:main_res_baseline} The main results of replacing the vocabulary of Pythia to Gemma. The best performance among the eight methods is displayed in \textbf{bold}. ${}^{*}$ indicates statistically significant improvements of 5\% level. ``+Align Rep.'' denotes the GloVe embeddings for tokens are converted into relative representations using 300 common tokens in both vocabularies before alignment following \cite{mosca-etal-2023-distinguishing}. 
}
\vspace{-4mm}
\end{table*}

\subsubsection{Cross-model Transfer}
\label{sec:tok_dist}
Unifying vocabulary with capable LLMs enables token-level distillation and transfers the knowledge learned into smaller models to decrease inference costs. 
In this section, training samples from downstream tasks and the corpus of Pile are used in the token-level distillation experiments. 
The logit of each token from the teacher model is taken as the soft label for Pythia to learn. 
Specifically, we add the KL-divergence loss between the logit from the teacher and student models to the original next token prediction loss on the training samples. 
The proportion of training samples is empirically set to 15\% to avoid a significant degradation in language modeling performance \citep{wei2023skywork}. 
There are two baseline methods: ``+ Direct tuning'', where models directly fine-tune on the training samples, and ``+ Sentence distill'' for comparison, where models fine-tune on the output text from the teacher model given the question as a prompt.

Table \ref{tab:token_distill_res} reports the results of two baseline methods and token-level distillation from three teacher models using 235M tokens. 
It can be found that token-level distillation is significantly better than sentence-level distillation. 
In the neural machine translation domain, token-level distillation outperforms sentence-level distillation when using larger student models, simpler texts, and abundant decoding information \citep{kim-rush-2016-sequence, wei2024sent}. 
Given the same teacher model Qwen2${}_{\text{7B}}$, the improvement of Pythia over the sentence-level distillation result reaches 4.4\%. 
The performance of Pythia${}_{\text{1B}}$ is even comparable with the vanilla Pythia${}_{\text{7B}}$ after token-level distillation. 
It is also noted that the knowledge transfer between models will be constrained in sentence-level distilling without unifying vocabulary, which further demonstrates the importance of unifying tokenizers between models.


\subsection{Vocabulary Adaptation Results}
\label{sec:vocab_adapt}
We show experimental results of replacing the Pythia vocabulary (50.3k) with the Gemma vocabulary (256.0k) using all methods in Table \ref{tab:main_res_baseline}. 
Given the same amount of tokens to fine-tune, it can be found that TokenAlign performs better than other baseline methods. 
The average improvement of TokenAlign over the strong baseline method ZeTT reaches 2.4\%, and 97.6\% performance of the vanilla model is reserved after vocabulary replacement. 
ZeTT requires more computation to train a hypernetwork for the parameters prediction, e.g., 661.2 GPU hours for Pythia${}_{\text{2.8B}}$, while our method only costs less than two hours on a CPU server with 128 cores to train GloVe embeddings and align tokens. 
Replace the corpus to train the GloVe embedding with 1B SlimPajama \citep{cerebras2023slimpajama} tokens brings comparable results (the ``w/ SlimPajama'' row). 
It demonstrates the robustness of our method on the pre-training corpus for token embedding and alignment matrix. 
Following \citet{moschella2023relative}, we also evaluate the method that converts token representations into relative ones using 300 common tokens in both vocabularies as anchors before calculating the alignment matrix $M_{s\to t}$, which brings better performance. 






\subsection{Analysis}

\begin{figure}[ht]
\centering
\includegraphics [width=0.42\textwidth]{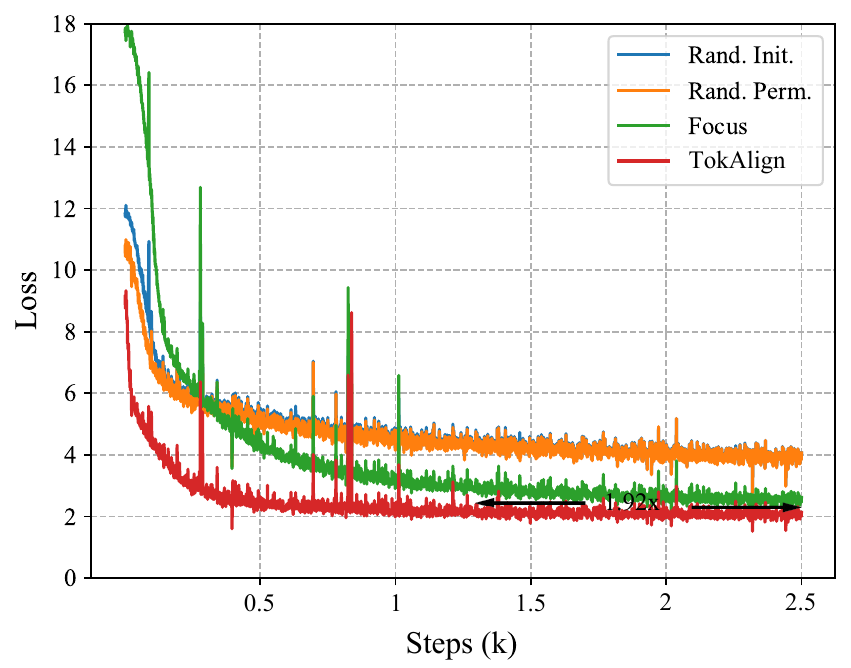}
\vspace{-2mm}
\caption{\label{fig:init_loss_5b} The training loss of Pythia${}_{\text{2.8B}}$.}
\vspace{-4mm}
\end{figure}

The loss curves of Pythia${}_{\text{2.8B}}$ with different methods during the first 2.5k steps are shown in Figure \ref{fig:init_loss_5b}. 
We find that TokAlign brings a better initialization and decreases the first-step training loss from 17.8 (Focus) to 9.5. 
Moreover, the training process with TokAlign is faster than other methods, which reaches 2.75 at the 1.3k step and is 1.92x (2.5/1.3) speed up than Focus. 

\begin{figure}[ht]
    \centering
    \subfigure[BLEU-1($\mathcal{C}_{t}$, $\mathcal{C}_{t}^{'}$)]{\includegraphics [scale=0.25]{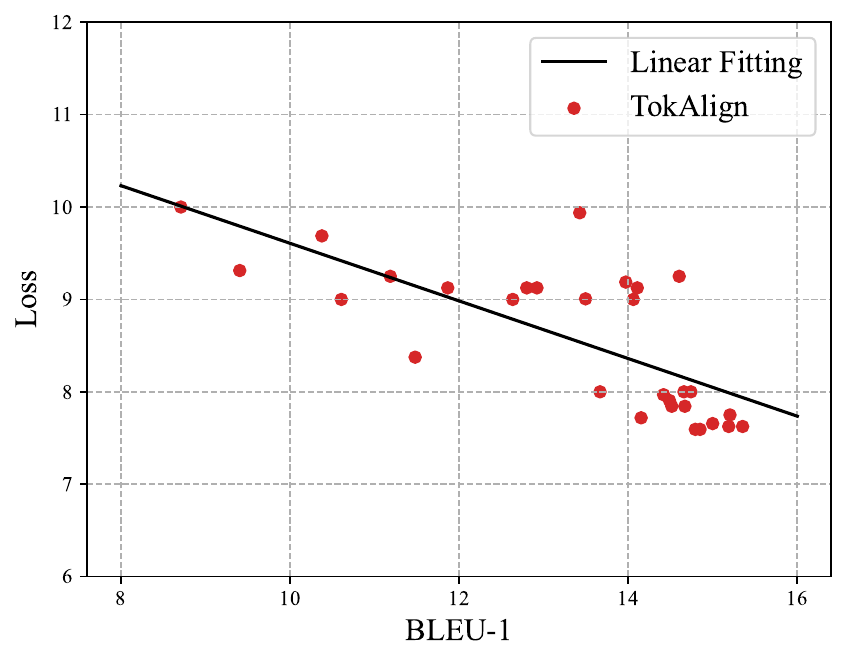}\label{fig:bleu1_loss}}
    \vspace{-1mm}
    \subfigure[BERTScore($\mathcal{C}$, $\mathcal{C}^{'}$)]{\includegraphics [scale=0.25]{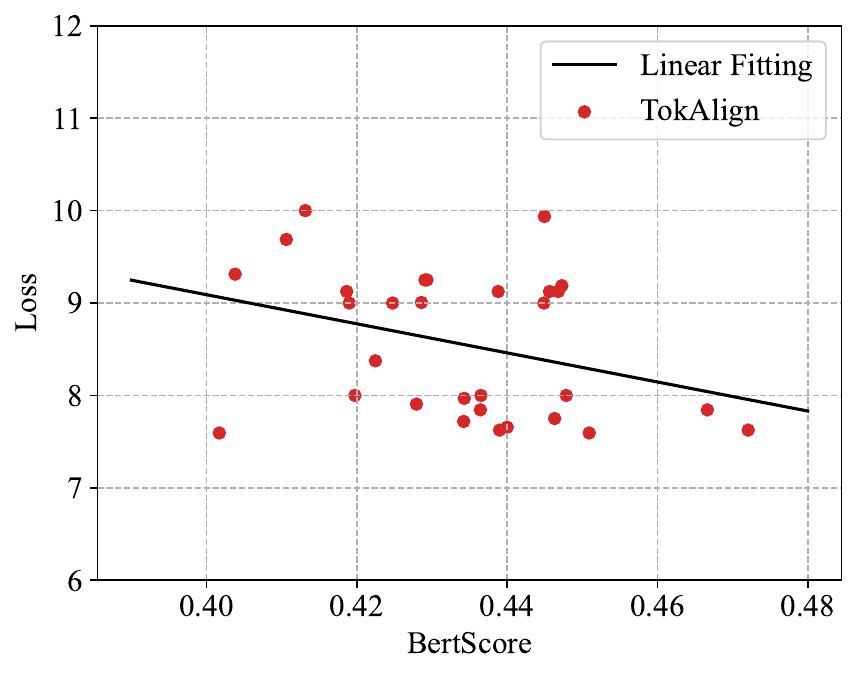}\label{fig:bertscore_loss}}
    \vspace{-1mm}
    \caption{\label{fig:loss_fit}The relationship between initial training loss and BLEU-1 (a) or BERTScore (b) for Pythia$_{\text{1B}}$.}
    \vspace{-4mm}
\end{figure}

\begin{table*}[thp]

\renewcommand\arraystretch{0.85}

\centering
\scriptsize

\setlength{\tabcolsep}{1.4mm}

 \begin{tabu}{lr|cc|cc|cc|cc|cc|cc|cc}
 
 \toprule[1.2pt]
  
  \multicolumn{2}{c}{ } & \multicolumn{2}{c}{\textbf{ARC-E}} & \multicolumn{2}{c}{\textbf{BoolQ}} & \multicolumn{2}{c}{\textbf{HellaSwag}} & \multicolumn{2}{c}{\textbf{OpenbookQA}} & \multicolumn{2}{c}{\textbf{PIQA}}  & \multicolumn{2}{c}{\textbf{WinoGrande}} & \multicolumn{2}{c}{\textbf{Avg}} \\

  \cmidrule(r){3-4} \cmidrule(r){5-6} \cmidrule(r){6-8} \cmidrule(r){9-10} \cmidrule(r){11-12} \cmidrule(r){13-14} \cmidrule(r){15-16} \noalign{\smallskip}

 \multicolumn{1}{c}{\textbf{Model}} & \multicolumn{1}{c}{$\#$\textbf{$\mathcal{V}$ (k)}} & \textbf{0} & \textbf{5} & \textbf{0} & \textbf{5} & \textbf{0} & \textbf{5} & \textbf{0} & \textbf{5} & \textbf{0} & \textbf{5} & \textbf{0} & \textbf{5} & \textbf{0} &  \multicolumn{1}{c}{\textbf{5}} \\

\midrule[0.8pt]

$\text{Pythia}_{\text{1B}}$
&$50.3$&$56.82$&$58.71$&$60.43$&$57.37$&$37.68$&$37.66$&$18.80$&$19.00$&$70.40$&$71.49$&$53.20$&$52.01$&$49.55$&$49.37$\\

   \cdashlinelr{1-16}

$\text{\ \ \ $\to$\ }{\text{Gemma}}$
&$256.0$&$54.46$&$56.86$&$\textbf{58.90}$&$52.26$&$36.16$&$36.27$&$\textbf{21.00}$&$20.20$&$67.74$&$68.50$&$52.25$&$50.91$&$48.42$&$47.50$\\

$\text{\ \ \ $\to$\ }{\text{Qwen2}}$
&$152.1$&$54.46$&$57.07$&$54.80$&$49.79$&$37.18$&$37.04$&$19.20$&$18.40$&$68.44$&$\textbf{70.24}$&$53.35$&$52.80$&$47.91$&$47.56$\\

$\text{\ \ \ $\to$\ }{\text{LLaMA2}}$
&$32.0$&$49.45$&$52.02$&$58.32$&$55.75$&$35.38$&$35.45$&$18.80$&$17.80$&$66.32$&$66.65$&$53.91$&$50.91$&$47.03$&$46.43$\\

$\text{\ \ \ $\to$\ }{\text{LLaMA3}}$
&$128.0$&$\textbf{54.63}$&$\textbf{57.28}$&$55.84$&$\textbf{53.70}$&$\textbf{37.34}$&$\textbf{37.43}$&$20.20$&$\textbf{20.40}$&$\textbf{69.04}$&$70.18$&$\textbf{54.46}$&$\textbf{53.43}$&$\textbf{48.59}$&$\textbf{48.74}$\\

\midrule[0.8pt]

 $\text{Pythia}_{\text{2.8B}}$
&$50.3$&$63.80$&$67.00$&$63.91$&$65.14$&$45.32$&$45.04$&$24.00$&$25.20$&$74.05$&$74.43$&$58.64$&$60.77$&$54.95$&$56.26$\\

   \cdashlinelr{1-16}

$\text{\ \ \ $\to$\ }{\text{Gemma}}$
&$256.0$&$61.62$&$65.15$&$63.82$&$\textbf{65.47}$&$43.13$&$43.18$&$23.40$&$\textbf{25.80}$&$72.14$&$72.42$&$58.17$&$\textbf{61.17}$&$53.71$&$\textbf{55.53}$\\

$\text{\ \ \ $\to$\ }{\text{Qwen2}}$
&$152.1$&$\textbf{62.54}$&$\textbf{66.04}$&$62.35$&$63.55$&$44.46$&$44.39$&$23.20$&$24.60$&$\textbf{73.50}$&$\textbf{73.56}$&$\textbf{59.04}$&$59.59$&$54.18$&$55.29$\\

$\text{\ \ \ $\to$\ }{\text{LLaMA3}}$
&$128.0$&$61.83$&$64.60$&$\textbf{64.40}$&$63.94$&$\textbf{44.62}$&$\textbf{44.59}$&$\textbf{23.80}$&$25.60$&$73.45$&$73.29$&$57.54$&$58.72$&$\textbf{54.27}$&$55.12$\\

\midrule[0.8pt]

 $\text{Pythia}_{\text{6.9B}}$
&$50.3$&$65.99$&$69.23$&$62.84$&$62.02$&$47.56$&$47.64$&$25.00$&$27.00$&$74.65$&$75.41$&$60.46$&$62.43$&$56.08$&$57.29$\\

   \cdashlinelr{1-16}

$\text{\ \ \ $\to$\ }{\text{Gemma}}$
&$256.0$&$65.40$&$68.35$&$62.39$&$59.57$&$45.75$&$45.86$&$22.00$&$25.60$&$73.39$&$74.10$&$60.38$&$61.17$&$54.89$&$55.77$\\

$\text{\ \ \ $\to$\ }{\text{Qwen2}}$
&$152.1$&$65.57$&$\textbf{68.43}$&$\textbf{64.07}$&$57.61$&$46.84$&$46.91$&$\textbf{25.60}$&$25.40$&$73.45$&$74.65$&$61.17$&$63.14$&$56.12$&$56.02$\\

$\text{\ \ \ $\to$\ }{\text{LLaMA3}}$
&$128.0$&$\textbf{66.46}$&$68.35$&$63.79$&$\textbf{60.64}$&$\textbf{47.28}$&$\textbf{47.31}$&$\textbf{25.60}$&$\textbf{28.20}$&$\textbf{74.48}$&$\textbf{75.84}$&$\textbf{61.48}$&$\textbf{63.30}$&$\textbf{56.52}$&$\textbf{57.27}$\\

\bottomrule[1.2pt]
\end{tabu}
\vspace{-1mm}

\caption{\label{tab:main_res} The benchmark results of replacing different tokenizers using TokAlign. The overlapping ratio between the vocabulary of Pythia and other models are 6.23\% (Gemma), 26.92\% (Qwen2), 28.10\% (LLaMA2), 32.85\% (LLaMA3). 
}

\vspace{-2mm}
\end{table*}

\begin{figure*}[th]
    \centering
    \subfigure[Learning rate $\text{2e}^{\text{-5}}$]{\includegraphics [scale=0.32]{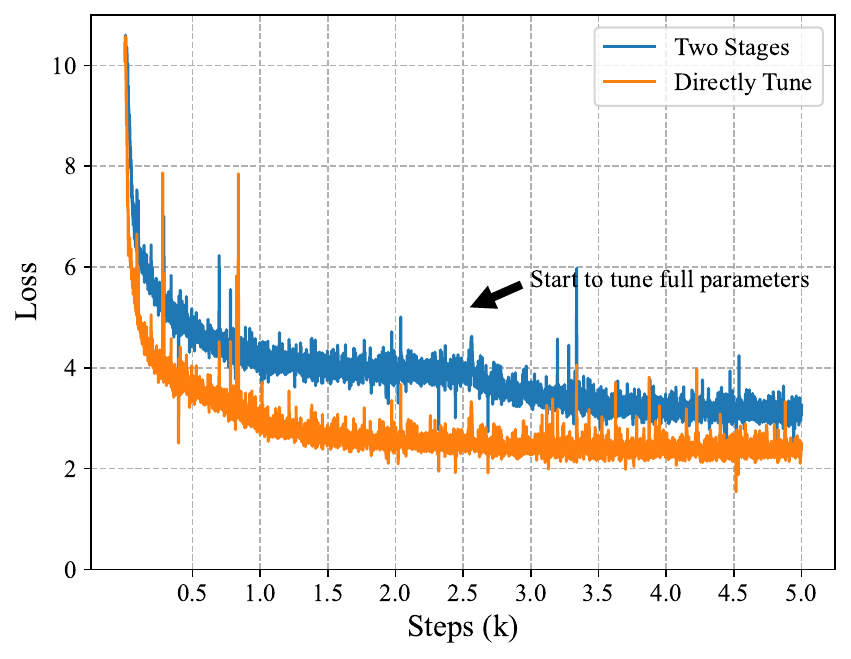}\label{fig:lr2e-5}}
    \subfigure[Learning rate $\text{8e}^{\text{-5}}$]{\includegraphics [scale=0.32]{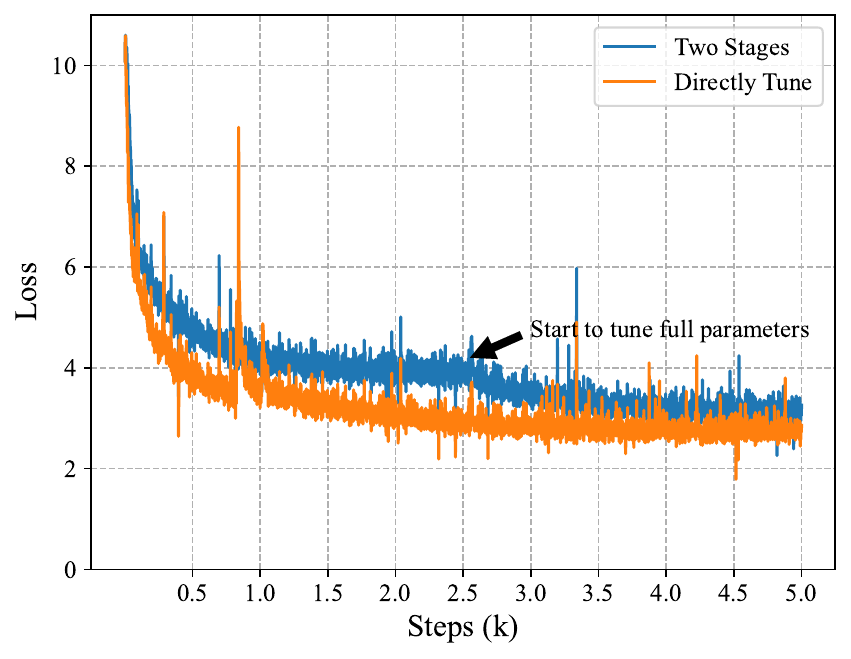}\label{fig:lr8e-5}}
    \subfigure[Learning rate $\text{6.4e}^{\text{-4}}$]{\includegraphics [scale=0.32]{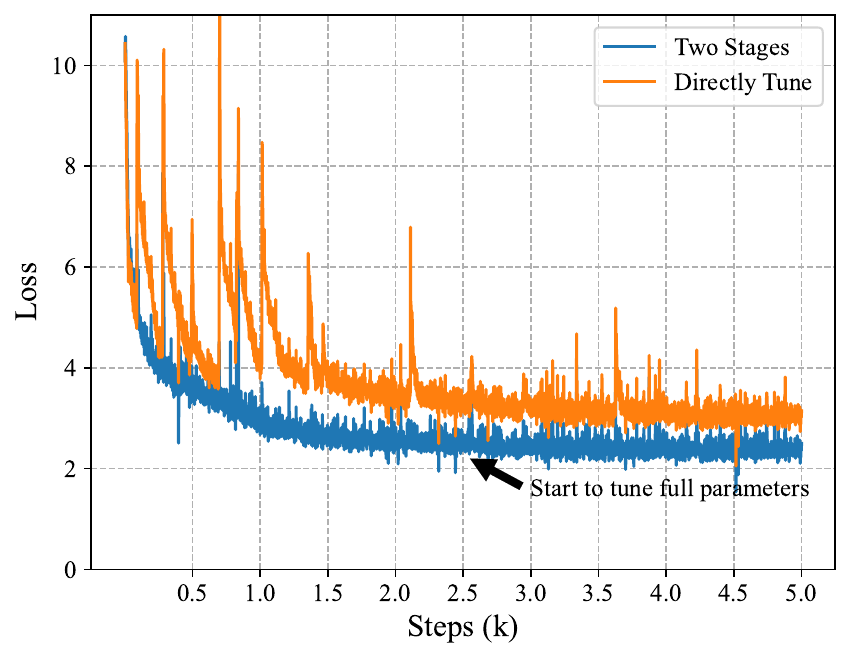}\label{fig:lr6.4e-4}}
    \vspace{-2mm}
    \caption{\label{fig:loss_2stage}The loss curve of Pythia${}_{\text{1B}}$ under two-stage tuning or direct full parameters tuning.}
    \vspace{-2mm}
\end{figure*}

\paragraph{Better alignment brings better initialization.}
We further investigate the impact of the learned alignment matrix $M_{s\to t}$ by changing the hyper-parameters of GloVe. 
It is noted that different alignment matrices $M_{s\to t}$ bring different initial parameters, and also result in different BLEU-1 scores on the same evaluation corpus. 
Figure \ref{fig:bleu1_loss} illustrates the negative relationship between the first-step training loss and BLEU-1. 
The sentence embedding model named ``all-mpnet-base-v2'' \citep{song2020mpnet} is adopted in the BERTScore evaluation. 
As shown in Figure \ref{fig:bertscore_loss}, it also shows a clear negative relationship with the initial training loss.
In other words, the higher the BLEU-1 score or BERTScore for the alignment matrix $M_{s\to t}$, the better the initial parameter is.  



\paragraph{More overlapping comes to faster convergence and higher performance.} TokAlign is further applied to the other three target tokenizers: Qwen2, LLaMA2, and LLaMA3. 
Table \ref{tab:main_res} reports the performance of models after replacing vocabulary on six datasets. 
TokAlign recovers 98.0\% performance of the base model on average with only 5k steps. 
Given a target vocabulary with more tokens than the one of Pythia (50.3k), it can be found that a higher overlapping ratio brings a better performance of model replaced (97.6\% for Gemma to 99.1\% for LLaMA3). 
The zero-shot in-context learning results for Pythia${}_{\text{6.9B}}$ with LLaMA3 vocabulary even surpass the vanilla base model. 
The results of Pythia${}_{\text{1B}}$ with LLaMA2 vocabulary are only 94.5\%, which is inferior to the average result. 
We argue that it may come from the missing 75.0M parameters (7.4\% for Pythia${}_{\text{1B}}$) after switching to a 32.0k vocabulary from the 50.3k vocabulary.

Figure \ref{fig:1b2other_loss} in Appendix \ref{appendix:convergence} shows the training loss curve. 
The replacing process of the Gemma tokenizer is the slowest, which may come from the only 6.23\% overlapping ratio between two vocabularies. 
It is in line with the result of random initialization in Figure \ref{fig:1b2other_rand_init}. 
Appendix \ref{appendix:convergence} reports more quantitative results by shuffling the alignment matrix, which further demonstrates the importance of token alignment.


\paragraph{Two-stage tuning brings a more stable convergence.} To replace the tokenizer and keep the performance of the vanilla model, we only fine-tune the vocabulary-related parameters at the first stage. 
The main reason for two-stage tuning is to take these parameters as the adapters of different tokenizers and avoid the well-trained parameters of the internal layer being distracted by the new initialized parameters. 

\begin{figure}[ht]
    \centering
    \includegraphics [scale=0.45]{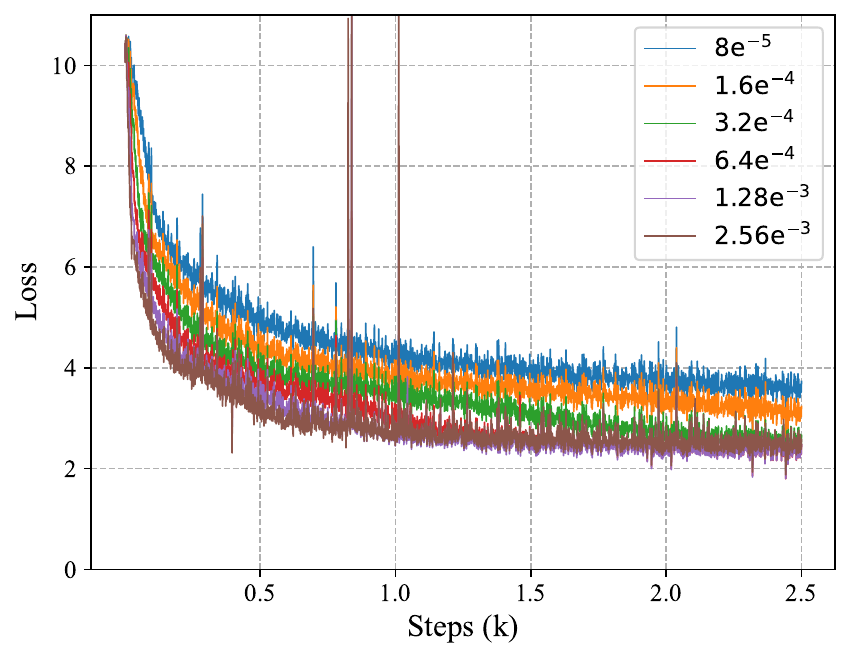}
    \vspace{-2mm}
    \caption{\label{fig:1b2gemma_lr}The training loss curve of Pythia${}_{\text{1B}}$ for learning rate used during replacing to the Gemma tokenizer.}
    \vspace{-4mm}
\end{figure}

Figure \ref{fig:loss_2stage} illustrates that our two-stage tuning method makes the convergence more stable under a high learning rate like $\text{6.4e}^{\text{-4}}$, which comes to better performance after vocabulary adaptation. 
It is noted that the loss spike also occurs at the first stage, fine-tuning vocabulary-related parameters only, under such a high learning rate like $\text{2.56e}^{\text{-3}}$ in Figure \ref{fig:1b2gemma_lr}. 

\section{Conclusion and Future Work}
In this paper, we introduce a method named TokAlign to replace the tokenizer of large language models from a token-token co-occurrence view. 
Extensive experiments demonstrate that TokAlign restores the performance of vanilla models after vocabulary adaptation, which enables cross-lingual knowledge transfer and deep knowledge transfer between models like token-level distillation. 

Beyond replacing the vocabulary of large language models, our method can be extended to replace the vocabulary of multi-modal models by aligning different modal tokens. 
The other direction is to develop a faster method, e.g., incorporating meta-learning in the two-stage tuning method to speed up the convergence. 

\section*{Limitations}
The first limitation comes from the assumption that the pre-training data distribution is available. 
We conduct experiments on Pythia with different parameter amounts, which provide public model weights and pre-training corpus. 
Due to the limited computation resource budget, open-source language models with unknown pre-training corpus like Mistral \citep{jiang2023mistral} are not investigated in this work. 
However, the pre-training corpus distribution of open-weighted large language models can be roughly inferred by the BPE vocabulary \citep{hayase2024data}. 
It can re-construct a similar pre-training corpus to conduct replacing tokenizer experiments. 

Another limitation is the additional 5k steps for vocabulary adaptation to replace a tokenizer. 
From the loss curve of TokAlign (Figure \ref{fig:1b2other_loss}), we find that the start of full parameters tuning can be faster, which may result in a better balance between performance and computational budget. 
Appendix \ref{app:adapt_2b} reports a preliminary result with only 2k steps, where TokAlign also shows a promising result. 


\section*{Acknowledgements}
We would like to thank the anonymous reviewers for their helpful discussions and valuable comments. 
The research work was supported by the National Key R\&D Program of China (No. 2022ZD0160602) and the Strategic Priority Research Program of Chinese Academy of Sciences (No. XDA04080400).

\bibliography{anthology, custom}

\appendix

\begin{figure*}[ht]
    \centering
    \subfigure[Vocabulary coverage]{\includegraphics [scale=0.45]{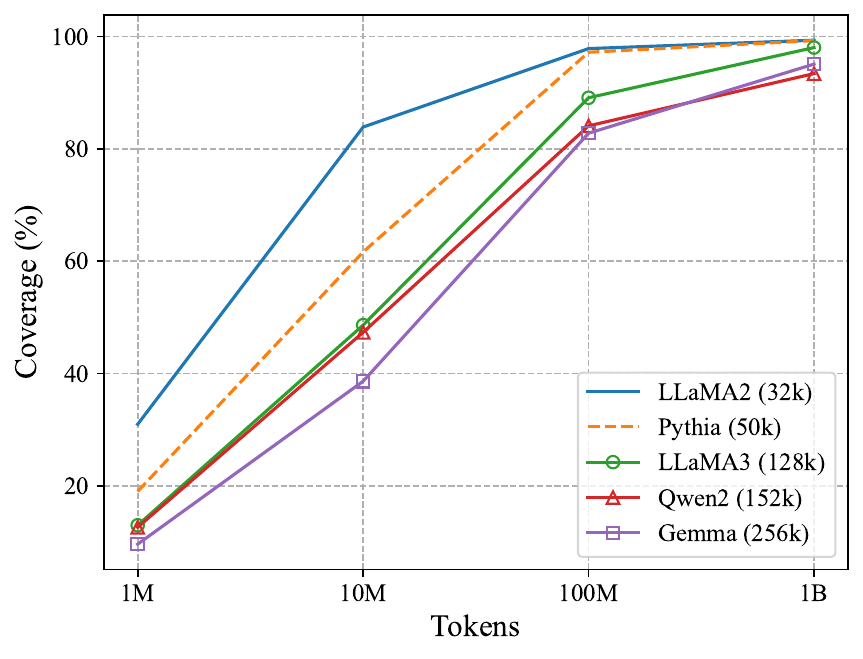}\label{fig:1b2glove_vec}}
    \subfigure[Initial loss with Gemma tokenizer]{\includegraphics [scale=0.45]{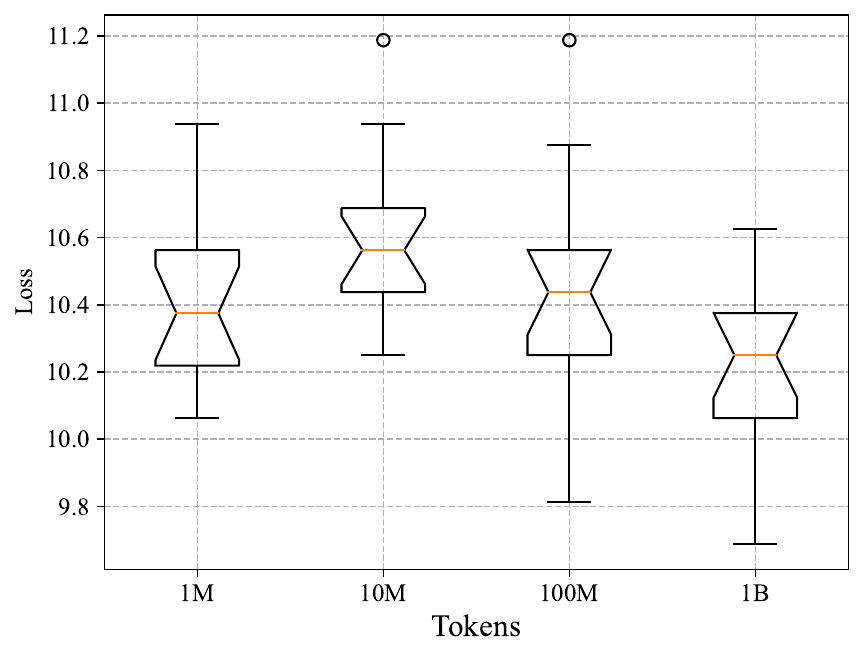}\label{fig:1b2glove_loss}}
    \vspace{-2mm}
    \caption{\label{fig:glove_hyper}The average vocabulary coverage (a) and initial training loss of Pythia${}_{\text{1B}}$ (b) under different amount tokens to train the GloVe vector.}
    \vspace{-4mm}
\end{figure*}


\section{Hyper-parameters}
\label{appendix:param}
\paragraph{GloVe Training}
We empirically train GloVe vectors with 1B tokens, which covers most tokens from Gemma (95.10\%), Qwen2 (93.40\%), LLaMA2 (99.35\%), and LLaMA3 (98.04\%). 
The dimension size is set to 300. 
The max training iteration and the size of the slide window are 15. 

\paragraph{Model Tuning}
The optimizer adopted in this work is AdamW \citep{loshchilov2019adamw}, where $\beta_1 = 0.9$ and $\beta_2 = 0.999$. 
The learning rate for baseline methods is set to 5e-5 to reduce the loss spike in Figure \ref{fig:lr8e-5} and Figure \ref{fig:lr6.4e-4}. 
We adopt bf16 mixed precision training, ZeRO-1, and flash-attention to save GPU memory cost and speed up the training process \citep{micikevicius2018mixed, rasley2020deepspeed, dao2022flash}. 
Following \citet{biderman2023pythia}, the batch size is set to 2M tokens and the max sequence length is 2048.

\section{Additional Results}
\label{appendix:res}

\subsection{Tokenizer Compression Rate}
\label{appendix:tok}
Table \ref{tab:tok} reports detailed compression rates of tokenizers across different domains and languages. 
We randomly sample 10 subsets or languages from vanilla datasets \citep{azerbayev2024llemma, kocetkov2023stack} to estimate the compression rate. 
Following \citet{lai-etal-2023-okapi}, the division of languages between ``\textbf{High}'', ``\textbf{Medium}'' and ``\textbf{Low}'' is determined by the available amount resource on CommonCrawl. 

\begin{table*}[htp]

\renewcommand\arraystretch{1.2}

\centering
\small

\setlength{\tabcolsep}{2.2mm}

 \begin{tabu}{c|c|ccccc}
 
 \toprule[1.2pt]
  
  \multicolumn{2}{c}{ } & \multicolumn{5}{c}{\textbf{Tokenizer}} \\

  \cmidrule(r){3-7} \noalign{\smallskip}

 \multicolumn{1}{c}{\textbf{Domain}} & \multicolumn{1}{c}{\textbf{Subset / Language}} & \multicolumn{1}{c}{\textbf{Gemma}} & \textbf{LLaMA3} & \textbf{LLaMA2} & \textbf{Qwen2} & \textbf{Pythia} \\

\midrule[0.8pt]
\multirow{5}{*}{\makecell{\textbf{Math} \\ \citep{azerbayev2024llemma}}} &
$ArXiv$&$2.8561$&$2.7765$&$2.7040$&$2.7445$&$2.8489$ \\
&$Textbooks$&$4.0883$&$4.3270$&$3.6500$&$4.2899$&$3.9464$ \\
&$Wikipedia$&$3.1753$&$3.2049$&$2.8792$&$3.0312$&$3.2898$ \\
&$ProofWiki$&$2.7538$&$2.8115$&$2.5996$&$2.7900$&$2.7363$ \\
&$StackExchange$&$3.2062$&$3.2814$&$3.0094$&$3.2107$&$3.2222$ \\
&$WebPages$&$3.9885$&$4.0655$&$3.5070$&$3.8720$&$4.1136$ \\

\cdashlinelr{1-7}

\multirow{10}{*}{\makecell{\textbf{Code} \\ \citep{kocetkov2023stack}}} &
$Python$&$3.3401$&$4.1331$&$3.0072$&$4.0339$&$3.2328 $ \\
&$Java$&$3.7175$&$4.4900$&$3.2193$&$4.4141$&$3.4914 $ \\
&$Go$&$2.9274$&$3.4797$&$2.5189$&$3.3870$&$2.8542 $ \\
&$VHDL$&$2.1038$&$2.4814$&$1.8724$&$2.2961$&$2.1395 $ \\
&$ActionScript$&$3.3470$&$3.9717$&$2.7852$&$3.9180$&$3.2949 $ \\
&$Scheme$&$2.7178$&$3.3045$&$2.4586$&$2.9713$&$2.9326 $ \\
&$Haml$&$3.2423$&$3.8429$&$2.9588$&$3.8002$&$3.1016 $ \\
&$Xbase$&$2.8739$&$3.4325$&$2.3300$&$3.3475$&$2.7837 $ \\
&$Mako$&$3.4387$&$4.0746$&$3.1238$&$4.0311$&$3.2844 $ \\
&$EmberScript$&$1.4104$&$1.9017$&$1.3819$&$1.4082$&$2.1540 $ \\

\cdashlinelr{1-7}
\multirow{10}{*}{\makecell{\textbf{High-Langs} \\ \citep{nguyen2023culturax}}} &
$English$&$4.4971 $&$4.6042 $&$3.8647 $&$4.4875 $&$4.4505 $ \\
&$Russian$&$6.7529 $&$5.8131 $&$4.9275 $&$5.3559 $&$3.5802 $ \\
&$Spanish$&$4.6068 $&$3.8416 $&$3.4517 $&$3.8330 $&$3.3655 $ \\
&$German$&$4.4605 $&$3.6314 $&$3.4417 $&$3.6041 $&$3.1096 $ \\
&$French$&$4.2258 $&$3.7378 $&$3.4445 $&$3.7243 $&$3.3565 $ \\
&$Chinese$&$3.7378 $&$3.2373 $&$1.8434 $&$3.9859 $&$1.9896 $ \\
&$Italian$&$4.2211 $&$3.4952 $&$3.3320 $&$3.4573 $&$3.1928 $ \\
&$Portuguese$&$4.2731 $&$3.6030 $&$3.2031 $&$3.5850 $&$3.2022$ \\ 
&$Polish$&$3.5583 $&$2.8548 $&$2.6639 $&$2.9464 $&$2.4333 $ \\
&$Japanese$&$5.7640 $&$4.2796 $&$2.4701 $&$4.7059 $&$2.9326 $ \\

\cdashlinelr{1-7}
\multirow{10}{*}{\makecell{\textbf{Medium-Langs}\\ \citep{nguyen2023culturax}}} &
$Czech$&$3.3402 $&$3.2875 $&$2.5978 $&$2.4490 $&$2.3884 $ \\
&$Vietnamese$&$4.5376 $&$4.2766 $&$1.9699 $&$4.2877 $&$2.0382 $ \\
&$Persian$&$5.6465 $&$5.3015 $&$1.7938 $&$3.1923 $&$2.3707 $ \\
&$Hungarian$&$3.2337 $&$2.6008 $&$2.6311 $&$2.5500 $&$2.3878 $ \\
&$Greek$&$4.4691 $&$4.5671 $&$1.8544 $&$2.1225 $&$3.0283 $ \\
&$Romanian$&$3.5558 $&$3.0566 $&$2.8355 $&$3.0083 $&$2.8981 $ \\
&$Swedish$&$3.7087 $&$3.1398 $&$2.9214 $&$3.0977 $&$2.9620 $ \\
&$Ukrainian$&$5.5141 $&$5.5985 $&$4.5904 $&$3.6179 $&$3.0702 $ \\
&$Finnish$&$3.2659 $&$2.6748 $&$2.4176 $&$2.6473 $&$2.6112 $ \\
&$Korean$&$3.3556 $&$3.6957 $&$1.5977 $&$3.3330 $&$1.5667 $ \\

\cdashlinelr{1-7}

\multirow{10}{*}{\makecell{\textbf{Low-Langs}\\ \citep{nguyen2023culturax}}} &
$Hebrew$&$4.0487 $&$1.8592 $&$1.7875 $&$4.3773 $&$2.0380 $ \\
&$Serbian$&$4.8596 $&$3.9234 $&$4.2642 $&$3.6267 $&$2.9896 $ \\
&$Tamil$&$5.6161 $&$2.0279 $&$2.2615 $&$2.4759 $&$1.9765 $ \\
&$Albanian$&$2.8919 $&$2.6536 $&$2.2945 $&$2.6037 $&$2.3631 $ \\
&$Azerbaijani$&$2.8585 $&$2.4857 $&$2.0407 $&$2.3797 $&$2.1534 $ \\
&$Kazakh$&$3.8172 $&$2.9176 $&$3.0869 $&$2.9263 $&$2.3236 $ \\
&$Urdu$&$4.4364 $&$2.8462 $&$1.7260 $&$2.7174 $&$1.9458 $ \\
&$Georgian$&$3.8237 $&$1.4828 $&$2.5595 $&$2.6951 $&$2.2077 $ \\
&$Armenian$&$3.2133 $&$1.1658 $&$1.7000 $&$1.8531 $&$1.3922 $ \\
&$Icelandic$&$2.7964 $&$2.4860 $&$2.3050 $&$2.4330 $&$2.3185 $ \\

\bottomrule[1.2pt]
\end{tabu}
\vspace{-1mm}

\caption{\label{tab:tok} The compression rates (bytes/token) of different tokenizers.
}

\vspace{-1mm}

\end{table*}


\subsection{GloVe Vectors}
\label{appendix:glove}
We show the effects of different token amounts for the GloVe vectors training in Figure \ref{fig:glove_hyper}. 
It can be found that 1B tokens used in this work provide a high vocabulary coverage ($>$90\%) and better initialization for Pythia${}_{\text{1B}}$. 
Due to the limited computation budget, experiments with more than 1B tokens are not conducted.

\begin{figure}[ht]
    \centering
    \includegraphics [scale=0.45]{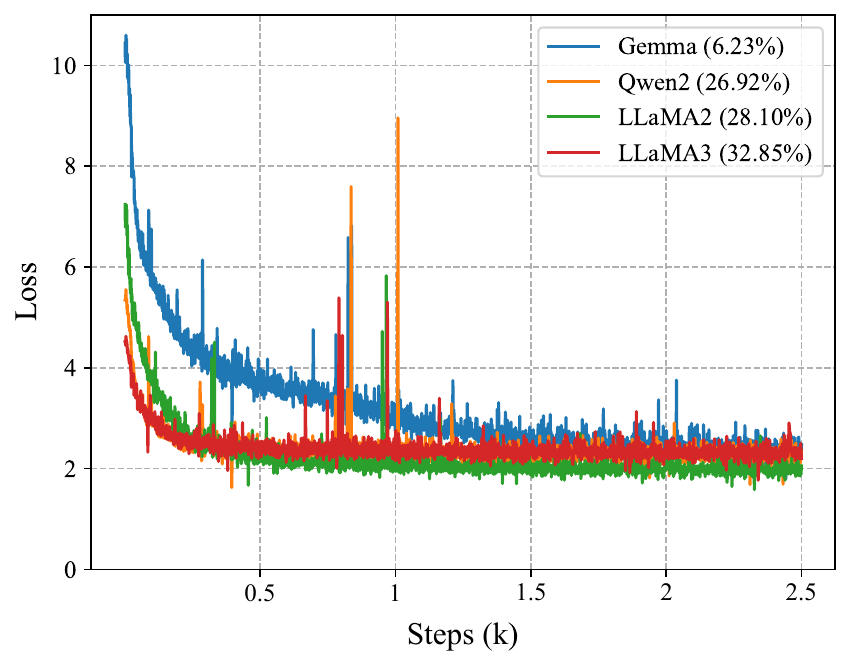}
    \vspace{-2mm}
    \caption{\label{fig:1b2other_loss}The training loss curve of Pythia${}_{\text{1B}}$ for different overlapping ratios.}
    \vspace{-4mm}
\end{figure}

\begin{figure}[ht]
    \centering
    \includegraphics [scale=0.45]{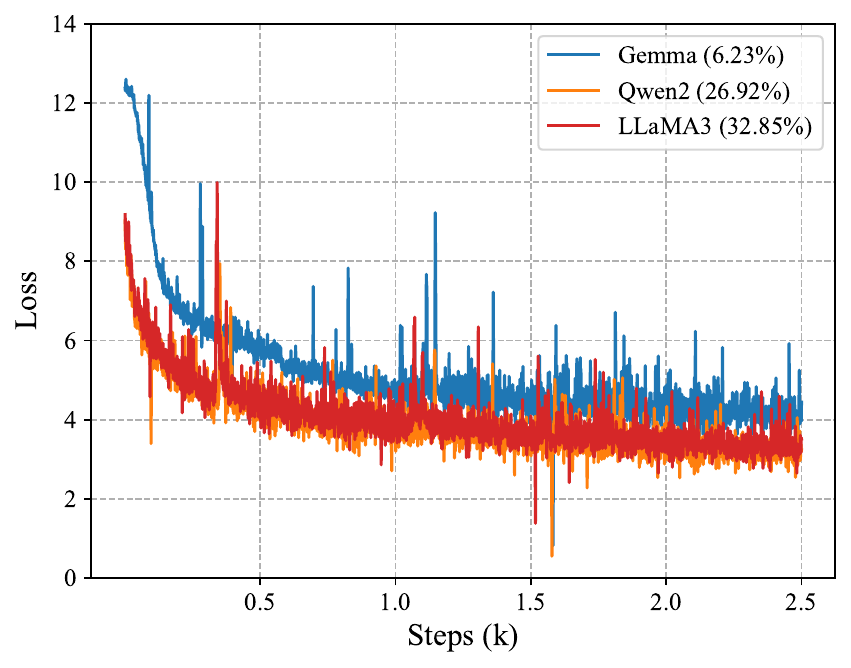}
    \vspace{-2mm}
    \caption{\label{fig:1b2other_rand_init}The training loss to different tokenizers using random initialization baseline.}
    \vspace{-4mm}
\end{figure}

\begin{table*}[thp]

\renewcommand\arraystretch{1.3}

\centering
\scriptsize

\setlength{\tabcolsep}{1.4mm}

\vspace{1mm}

 \begin{tabu}{lr|cc|cc|cc|cc|cc|cc|cc}
 
 \toprule[1.2pt]
  
  \multicolumn{2}{c}{ } & \multicolumn{2}{c}{\textbf{ARC-E}} & \multicolumn{2}{c}{\textbf{BoolQ}} & \multicolumn{2}{c}{\textbf{HellaSwag}} & \multicolumn{2}{c}{\textbf{OpenbookQA}} & \multicolumn{2}{c}{\textbf{PIQA}}  & \multicolumn{2}{c}{\textbf{WinoGrande}} & \multicolumn{2}{c}{\textbf{Avg}} \\

  \cmidrule(r){3-4} \cmidrule(r){5-6} \cmidrule(r){6-8} \cmidrule(r){9-10} \cmidrule(r){11-12} \cmidrule(r){13-14} \cmidrule(r){15-16} \noalign{\smallskip}

 \multicolumn{1}{c}{\textbf{Model}} & \multicolumn{1}{c}{$\#$\textbf{$\mathcal{V}$ (k)}} & \textbf{0} & \textbf{5} & \textbf{0} & \textbf{5} & \textbf{0} & \textbf{5} & \textbf{0} & \textbf{5} & \textbf{0} & \textbf{5} & \textbf{0} & \textbf{5} & \textbf{0} &  \multicolumn{1}{c}{\textbf{5}} \\

\midrule[0.8pt]

$\text{Pythia}_{\text{1B}}$
&$50.3$&$56.82$&$58.71$&$60.43$&$57.37$&$37.68$&$37.66$&$18.80$&$19.00$&$70.40$&$71.49$&$53.20$&$52.01$&$49.55$&$49.37$\\

   \cdashlinelr{1-16}

$\text{\ \ \ $\to$\ }{\text{Gemma}}$
&$256.0$&$51.09$&$52.44$&$53.12$&$52.35$&$35.00$&$35.05$&$20.20$&$18.60$&$64.80$&$65.83$&$53.12$&$51.62$&$46.22$&$45.98$\\

$\text{\ \ \ $\to$\ }{\text{Qwen2}}$
&$152.1$&$53.41$&$55.47$&$53.52$&$55.81$&$36.12$&$36.38$&$20.80$&$18.00$&$68.50$&$68.88$&$54.38$&$52.80$&$47.79$&$47.89$\\

$\text{\ \ \ $\to$\ }{\text{LLaMA3}}$
&$128.0$&$51.73$&$55.09$&$59.05$&$55.08$&$36.42$&$36.52$&$19.40$&$19.60$&$67.68$&$68.34$&$53.43$&$53.75$&$47.95$&$48.06$\\

\bottomrule[1.2pt]
\end{tabu}
\vspace{-2mm}

\caption{\label{tab:2b_res_baseline} The main results of replacing the vocabulary of Pythia for TokAlign using 2B tokens from the Pile corpus.
}

\end{table*}

\begin{figure}[ht]
    \centering
    \includegraphics [scale=0.45]{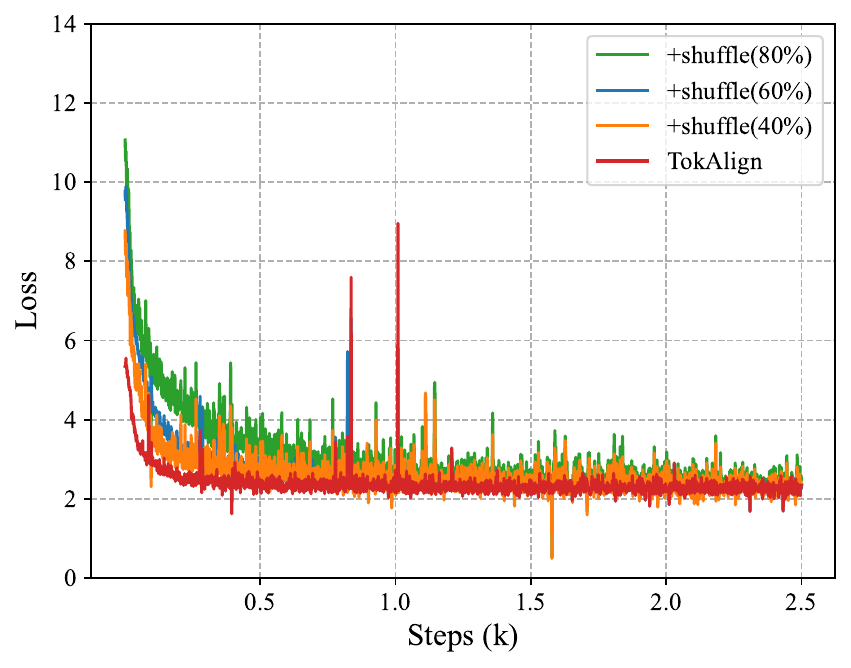}
    \vspace{-2mm}
    \caption{\label{fig:1b2qwen2}The training loss of Pythia${}_\text{1B}$ when replacing tokenizer to Qwen2 under different percentages of shuffling.}
    \vspace{-2mm}
\end{figure}


\subsection{Convergence Analysis}
\label{appendix:convergence}
To investigate the effect of overlapping rate between two tokenizers to the convergence of training, we plot Figure \ref{fig:1b2other_rand_init} for the random initialization baseline method. 
The convergence of Gemma tokenizer is slower than the other tokenizers and comes to worse results, which are similar to the case in Figure \ref{fig:1b2other_loss}. 

Moreover, we randomly shuffle the alignment matrix learned in TokAlign to imitate the case that other worse methods rather than cosine similarity to calculate the alignment matrix. 
Figure \ref{fig:1b2qwen2} shows that the higher percentage of randomly shuffle comes to higher initial training loss and slower convergence.


\subsection{Fast Vocabulary Adaptation Results}
\label{app:adapt_2b}

We further investigate a challenge condition that fine-tunes only 2B tokens to adapt the target vocabulary. 
To meet the requirement, we reduce the batch size to 1M tokens and set the number of fine-tuning steps to 2k. 
Table \ref{tab:2b_res_baseline} shows the results of adapting to the other 3 tokenizers using TokAlign. 
It can be found that 95.66\% performance of the vanilla model is recovered on average, which further demonstrates the effectiveness of our method.


\subsection{In-context Learning Results during Cross-lingual Transfer}
\label{appendix:cross_lingual_transfer}
Table \ref{tab:multilingual_0_shot} and \ref{tab:multilingual_5_shot} report the 0-shot and 5-shot in-context learning results on 4 multilingual datasets. 
The average improvement over the baseline method Focus is 2.35\% after language adaptation pre-training. 
We can find that the model initialized by TokAlign is comparable to the one of Focus after language adaptation pre-training, which mainly comes from the strong English performance preserved by TokAlign. 

\begin{table*}[thp]

\renewcommand\arraystretch{1.3}

\centering
\scriptsize

\setlength{\tabcolsep}{0.9mm}

 \begin{tabu}{l|ccccccc|ccccc|ccc|cccc|c}
 
 \toprule[1.2pt]
  
  \multicolumn{1}{c}{ } & \multicolumn{7}{c}{\textbf{XNLI}} & \multicolumn{5}{c}{\textbf{PAWS-X}} & \multicolumn{3}{c}{\textbf{XCOPA}}  & \multicolumn{4}{c}{\textbf{XStoryCloze}}  & \multicolumn{1}{c}{} \\

  \cmidrule(r){2-8} \cmidrule(r){9-13} \cmidrule(r){14-16} \cmidrule(r){17-20} \noalign{\smallskip}

 \multicolumn{1}{c}{\textbf{Model}} & \textbf{en} & \textbf{de} & \textbf{zh} & \textbf{ar} & \textbf{th} & \textbf{vi} & \textbf{ur} & \textbf{de} & \textbf{en} & \textbf{ja} & \textbf{ko} & \textbf{zh} & \textbf{th} & \textbf{vi} & \textbf{ta} & \textbf{en} & \textbf{zh} & \textbf{ar} & \multicolumn{1}{c}{\textbf{te}} & \multicolumn{1}{c}{\textbf{Avg}} \\

\midrule[0.8pt]

$\text{Pythia}_{\text{1B}}$&$46.2$&$38.6$&$\textbf{38.9}$&$\textbf{36.9}$&$35.2$&$\textbf{38.9}$&$34.9$&$48.9$&$48.3$&$52.9$&$\textbf{53.3}$&$54.1$&$53.4$&$52.6$&$55.4$&$\textbf{65.3}$&$48.6$&$48.2$&$52.2$&$\textbf{47.5}$\\

   \cdashlinelr{1-21}

$\text{\ \ \ w/\ }{\text{Focus Init.}}$&$32.8$&$32.2$&$33.6$&$33.6$&$33.5$&$32.0$&$32.8$&$44.8$&$46.0$&$48.9$&$44.8$&$44.7$&$51.4$&$47.6$&$\textbf{55.6}$&$45.9$&$48.6$&$\textbf{48.5}$&$46.8$&$42.3$\\

$\text{\ \ \ \ +\ LAT}$&$\textbf{47.0}$&$36.7$&$35.4$&$34.3$&$33.5$&$35.1$&$33.9$&$51.5$&$48.6$&$53.7$&$51.2$&$54.0$&$\textbf{54.4}$&$51.6$&$\textbf{55.6}$&$55.8$&$48.7$&$47.5$&$50.4$&$46.3$\\

$\text{\ \ \ w/\ }{\text{TokAlign Init.}}$
&$44.9$&$37.4$&$34.0$&$32.8$&$\textbf{35.3}$&$35.2$&$34.5$&$50.2$&$\textbf{50.3}$&$52.0$&$53.1$&$\textbf{54.4}$&$\textbf{54.4}$&$50.0$&$54.4$&$61.2$&$48.3$&$47.6$&$50.0$&$46.3$\\

$\text{\ \ \ \ +\ LAT}$&$44.4$&$\textbf{39.0}$&$38.7$&$35.6$&$35.1$&$37.8$&$\textbf{35.5}$&$\textbf{51.9}$&$49.3$&$\textbf{54.7}$&$53.1$&$50.6$&$54.2$&$\textbf{54.0}$&$52.8$&$64.7$&$\textbf{50.8}$&$48.0$&$\textbf{52.4}$&$\textbf{47.5}$\\

\midrule[0.8pt]

$\text{Pythia}_{\text{6.9B}}$&$\textbf{53.0}$&$40.7$&$\textbf{41.7}$&$\textbf{38.9}$&$37.3$&$41.3$&$35.1$&$49.4$&$47.1$&$52.9$&$52.2$&$52.4$&$\textbf{55.0}$&$53.6$&$53.6$&$\textbf{73.1}$&$\textbf{54.6}$&$\textbf{49.9}$&$\textbf{53.9}$&$49.2$\\

\cdashlinelr{1-21}

$\text{\ \ \ w/\ }{\text{Focus Init.}}$&$31.5$&$31.3$&$33.0$&$32.6$&$33.4$&$32.2$&$32.6$&$44.8$&$46.4$&$52.3$&$51.2$&$54.5$&$52.4$&$47.4$&$\textbf{56.0}$&$44.9$&$47.3$&$48.5$&$47.6$&$43.1$\\

$\text{\ \ \ \ +\ LAT}$&$45.1$&$37.7$&$35.3$&$33.4$&$35.0$&$38.1$&$33.8$&$49.5$&$49.0$&$52.6$&$54.5$&$55.3$&$52.0$&$51.2$&$53.8$&$61.5$&$48.3$&$47.3$&$53.4$&$46.7$\\

$\text{\ \ \ w/\ }{\text{TokAlign Init.}}$&$50.8$&$39.1$&$34.4$&$34.5$&$33.9$&$34.6$&$\textbf{35.2}$&$50.0$&$47.7$&$\textbf{53.9}$&$54.3$&$55.2$&$53.2$&$51.2$&$53.2$&$68.0$&$48.5$&$47.8$&$50.2$&$47.1$\\

$\text{\ \ \ \ +\ LAT}$&$49.2$&$\textbf{41.5}$&$37.8$&$36.9$&$\textbf{38.7}$&$\textbf{41.9}$&$34.7$&$\textbf{51.2}$&$\textbf{49.5}$&$53.5$&$\textbf{54.8}$&$\textbf{55.4}$&$53.4$&$\textbf{59.8}$&$52.8$&$73.0$&$53.9$&$49.2$&$53.6$&$\textbf{49.5}$\\

\bottomrule[1.2pt]
\end{tabu}
\vspace{-2mm}

\caption{\label{tab:multilingual_5_shot} Five-shot in-context learning results of cross-lingual transfer. 
}

\vspace{-5mm}

\end{table*}

\paragraph{Case study of multilingual token alignment.}
Table \ref{tab:multilingual_case} provides nine new tokens from three languages with their top 3 tokens in the source vocabulary for qualitative analyses. 
In most cases, a clear semantic relationship between two aligned tokens cannot be found. 
We argue that it may come from the following two reasons: 

\begin{table*}[htp]

\renewcommand\arraystretch{1.3}

\centering
\scriptsize

\setlength{\tabcolsep}{1.7mm}

 \begin{tabu}{l|ccc|ccc|ccc}
 
 \toprule[1.2pt]
  
  \multicolumn{1}{c}{ } & \multicolumn{3}{c}{\textbf{French}} & \multicolumn{3}{c}{\textbf{Chinese}} & \multicolumn{3}{c}{\textbf{Korean}} \\

  \cmidrule(r){2-4} \cmidrule(r){5-7} \cmidrule(r){8-10} \noalign{\smallskip}

 \multicolumn{1}{c}{\textbf{Top-3}} & \textbf{dire(speak)} & \textbf{aller(go)} & \textbf{oui(are)} & \textbf{\begin{CJK*}{UTF8}{gbsn}吃\end{CJK*}(eat)} & \textbf{\begin{CJK*}{UTF8}{gbsn}科学\end{CJK*}(science)} & \textbf{\begin{CJK*}{UTF8}{gbsn}智能\end{CJK*}(intelligence)} & \textbf{\begin{CJK}{UTF8}{mj}능\end{CJK}(competence)} & \textbf{\begin{CJK}{UTF8}{mj}집\end{CJK}(house)} & \textbf{\begin{CJK}{UTF8}{mj}왜\end{CJK}(why)} \\

 

\midrule[0.8pt]

\multicolumn{10}{c}{\textit{Qwen2 (Target Tokenizer)}} \\

\cdashlinelr{1-10}

$1$&ada&Ġsta&Ġsalv&allel&Ġantagon&$\_\{[$&Si&ĠBart&bst\\

$2$&ays&ĠÃ¨&Ġvas&Ġindicator&Ġign&liquid&uria&ĠPAT&rains \\

$3$&Ġ-&Ġdetermin&Ġexplos&Ġbasic&Ġcritic&Layer&ost&ĠEdgar&irc \\

\cdashlinelr{1-10}
   
\multicolumn{10}{c}{\textit{Gemma (Target Tokenizer)}} \\

\cdashlinelr{1-10}

$1$&Ġj&Cor&Tools&kernel&ĠLed&Ġcommittee&Ġmang&Ġcru&Ġcholesterol \\

$2$&Ġdar&Ġequality&directed&sentence&COUNT&ĠUND&ial&Ġcal&Ġmolecule \\

$3$&ba&Lex&afx&messages&Ġglycine&Ġfactors&Ġrebut&Ġmalt&apor \\

\bottomrule[1.2pt]
\end{tabu}
\vspace{-1mm}

\caption{\label{tab:multilingual_case} The case study of new tokens from other languages in the target vocabulary with top-3 source tokens aligned. 
The language family of French, Chinese, and Korean are Indo-European, Sino-Tibetan, and Koreanic, respectively. 
}

\vspace{-1mm}

\end{table*}

\begin{itemize} 
    \item BPE algorithm \citep{sennrich-etal-2016-neural} divides words into the sub-word units, also called tokens, from the statistical co-occurrence information. There may be less superficial semantic information in the tokens divided compared with words in the natural language. 
    
    \item The GloVe vector for each token is obtained from the token-token co-occurrence information. These aligned tokens often appear together, e.g., \begin{CJK*}{UTF8}{gbsn}科学\end{CJK*}(science) and ``Ġcritic'', \begin{CJK}{UTF8}{mj}왜\end{CJK}(why) and ``rains''.
\end{itemize}

Therefore, it is better to choose a matric to quantify the performance of the alignment matrix learned, for example, the BLEU-1 score or BERTScore in Section \ref{sec:align_eval}.

\begin{table*}[htp]

\renewcommand\arraystretch{1.3}

\centering
\scriptsize

\setlength{\tabcolsep}{4mm}


 \begin{tabu}{c|ccccrrr}
 
 \toprule[1.2pt]

\multicolumn{1}{c}{\textbf{Task}} & \multicolumn{1}{c}{\textbf{Dataset}}    &\multicolumn{1}{c}{\textbf{\#Lang}} & \multicolumn{1}{c}{\textbf{\#Class}} & \multicolumn{1}{c}{\textbf{Data Curation}} & \multicolumn{1}{r}{\textbf{\#Train}}  & \multicolumn{1}{r}{\textbf{\#Dev}}  &\multicolumn{1}{r}{\textbf{\#Test}}   \\

   \midrule[0.8pt]
Natural Language Inference          &$\text{XNLI}$                          & $15$  & $3$  &$\text{Translation}$   & $-$           & $2,490$       & $5,010$  \\
\cdashlinelr{1-7}
Paraphrase Detection                &$\text{PAWS-X}$                        & $7$   & $2$   &$\text{Aligned}$       & $-$           & $2,000$       & $2,000$ \\
\cdashlinelr{1-7}
\multirow{7}{*}{Reasoning}
                                    &$\text{ARC-Easy}$                      & $1$   & $4$   &$-$                    & $2,251$       & $570$         & $2,376$     \\
                                    &$\text{HellaSwag}$                     & $1$   & $4$   &$-$                    & $39,905$      & $10,042$      & $10,003$     \\
                                    &$\text{OpenbookQA}$                    & $1$   & $4$   &$-$                    & $4,957$       & $500$         & $500$     \\
                                    &$\text{PIQA}$                          & $1$   & $2$   &$-$                    & $16,000$      & $2,000$      & $3,000$     \\
                                    &$\text{XCOPA}$                         & $12$   & $2$  &$\text{Translation}$   & $33,810$      & $100$         & $500$     \\
                                    &$\text{XStoryCloze}$                   & $11$   & $2$  &$\text{Translation}$   & $361$         & $-$           & $1,511$    \\
                                    &$\text{WinoGrad}$                      & $1$   & $2$   &$-$                    & $40,398$           & $1,267$           & $1,767$    \\
\cdashlinelr{1-7}
Reading Comprehension               &$\text{BoolQ}$                          & $1$   & $2$  &$-$   & $9,427$           & $3,270$       & $-$  \\

\specialrule{0em}{0pt}{0pt}

\bottomrule[1.2pt]
\end{tabu}

\caption{\label{tab:datasets} Statistic of evaluation datasets used. 
}


\end{table*}


\newpage

\section{Evaluation Tasks}
\label{app:eval_tasks}
We report the statistics of evaluation tasks used in Table \ref{tab:datasets}. 
Here are the descriptions of these evaluation tasks: 
\paragraph{Natural Language Inference} aims to determine the semantic relationship (Entailment, neural, or contradiction) between the premise and hypothesis \citep{conneau-etal-2018-xnli}. 

\paragraph{Paraphrase Detection} requires the model to evaluate whether the second sentence is a paraphrase of the first sentence in this task \citep{yang-etal-2019-paws}. 

\paragraph{Commonsense Reasoning} is a task for the model to reason the gold answer based on the semantic coherence and physic rules \citep{clark2018arc, mihaylov-etal-2018-suit, zellers-etal-2019-hellaswag, ponti-etal-2020-xcopa, bisk-etal-2020-piqa, Sakaguchi-et-al-2020-winogrande, tikhonov-ryabinin-2021-heads}. 

\paragraph{Reading Comprehension} needs the model to infer whether the given passage can answer the query \citep{clark-etal-2019-boolq}. 

\section{Language Codes}
\label{appendix:lang_code}
We provide details of languages involved in Table \ref{tab:lang_codes}. 
Following \citet{lai-etal-2023-okapi}, languages are divided by the data ratios in CommomCrawl: High ($>$1\%), Medium ($>$0.1\%), and Low ($>$0.01\%). 

\begin{table}[htp]

\setlength{\tabcolsep}{4.7mm}
\centering
\small
\renewcommand\arraystretch{1.25}
\begin{center}
    \begin{tabular}{ccc}
        \toprule[1.2pt] 
        \multicolumn{1}{c}{\textbf{ISO 639-1}} & \multicolumn{1}{c}{\textbf{Language}}    &\multicolumn{1}{c}{\textbf{Family}} \\
                \midrule[0.8pt]
                AR      &Arabic      &Afro-Asiatic\\
                BN      &Bengali      &Indo-European\\
                DE      &German      &Indo-European\\
                EN      &English      &Indo-European\\
                JA      &Japanese      &Japonic\\
                KO      &Korean      &Koreanic\\
                TA      &Tamil      &Dravidian\\
                TE      &Telugu      &Dravidian\\
                TH      &Thai      &Kra-Dai\\
                UR      &Urdu      &Indo-European\\
                VI      &Vietnamese      &Austroasiatic\\
                ZH      &Chinese      &Sino-Tibetan\\
        \bottomrule[1.2pt]
    \end{tabular}
\end{center}
\vspace{-5mm}

\vspace{2mm}

\caption{\label{tab:lang_codes} Details of language codes in this work. 
}

\vspace{-2mm}

\end{table}

\section{Licenses of Scientific Artifacts}
We follow and report the licenses of scientific artifacts involved in Table \ref{tab:license}. 

\begin{table}[thp]
 \setlength{\tabcolsep}{1mm}
	\centering
	\small
	\renewcommand\arraystretch{1.25}
	\begin{center}
		\begin{tabular}{ll}
			\toprule[1.2pt]  
               \multicolumn{1}{c}{\textbf{Name}} & \multicolumn{1}{c}{\textbf{License}} \\
                \midrule[0.8pt]
                Transformers      & Apache 2.0 license     \\
                lm-evaluation-harness      & MIT license      \\
                matplotlib      & PSF license      \\
                Focus      & MIT license      \\
                WECHSEL      & MIT license      \\
                Pythia      & Apache 2.0 license      \\
                LLaMA3      & Meta LLaMA 3 community license      \\
                Qwen2      & Tongyi Qianwen license      \\
                Gemma      & Gemma license      \\
                The Pile      & MIT license      \\
			\bottomrule[1.2pt]
		\end{tabular}
	\end{center}
    \caption{\label{tab:license} Licenses of scientific artifacts involved in this work.}
\end{table}

\end{document}